\documentclass{ieeeaccess}
\usepackage{cite}
\usepackage{amsmath,amssymb,amsfonts}
\usepackage{algorithmic}
\usepackage{graphicx}
\usepackage{float}
\usepackage{kotex}
\usepackage{textcomp}
\usepackage{booktabs} 
\usepackage{tabularx}
\usepackage{pdflscape}
\usepackage{makecell}
\usepackage{multirow}
\usepackage{threeparttable}
\usepackage{hyperref}

\makeatletter
\@ifundefined{xfigwd}{\newdimen\xfigwd}{}
\makeatother

\usepackage[caption=false,font=footnotesize]{subfig}

\begin{document}
\history{Date of publication xxxx 00, 0000, date of current version xxxx 00, 0000.}
\doi{10.1109/ACCESS.2017.DOI}

\title{Benchmarking Visual Feature Representations for LiDAR-Inertial-Visual Odometry Under Challenging Conditions }
\author{%
  \uppercase{eunseon choi}\authorrefmark{1},
  \uppercase{junwoo hong}\authorrefmark{1}, 
  \uppercase{daehan lee}\authorrefmark{1}, 
  \uppercase{sanghyun park}\authorrefmark{1}, \newline
  \uppercase{hyunyoung jo}\authorrefmark{1}, 
  \uppercase{SunYoung Kim}\authorrefmark{2}, 
  \IEEEmembership{Member, IEEE},
  \uppercase{ChangHo Kang}\authorrefmark{3}, \newline 
  \uppercase{SeongSam KIM}\authorrefmark{4}, 
  \uppercase{YongHan JUNG}\authorrefmark{4}, 
  \uppercase{JungWook PARK}\authorrefmark{4},  \newline
  \uppercase{Seul KOO}\authorrefmark{4},
  \uppercase{and soohee han}\authorrefmark{1,5}, \IEEEmembership{Senior Member, IEEE}%
}
\address[1]{Department of Covergence IT Engineering, Pohang University of Science and Technology(POSTECH), Pohang 37673, Republic of Korea}
\address[2]{School of Mechanical
Engineering, Kunsan National University, Jeonbuk 54150, Republic of Korea}
\address[3]{Department of Artificial Intelligence and Robotics, Sejong University, Seoul 05006, Republic of Korea}
\address[4]{Disaster Scientific Investigation Division Disaster Investigation Technology Team, 
National Disaster Management Research Institute, 
Ministry of the Interior and Safety, Ulsan 44429, Republic of KOREA
}
\address[5]{Department of Electrical Engineering, POSTECH, Pohang 37673, Republic of Korea}

\tfootnote{This research was conducted as part of the project “Disaster Field
Investigation using Mobile Robot technology ($\mathrm{V}$
)”, which was
supported by the NDMI (National Disaster Management
Research Institute) under the project number NDMI-MA-2025-
06-01. The authors would like to acknowledge the financial
support of the NDMI.
}

\markboth
{E. Choi \headeretal: Benchmarking Visual Features for LiDAR-Inertial-Visual Odometry}
{E. Choi \headeretal: Benchmarking Visual Features for LiDAR-Inertial-Visual Odometry}

\corresp{Corresponding author: Soohee Han (e-mail: soohee.han@postech.ac.kr).}

\begin{abstract}
Accurate localization in autonomous driving is critical for successful missions including environmental mapping and survivor searches. In visually challenging environments, including low-light conditions, overexposure, illumination changes, and high parallax, the performance of conventional visual odometry methods significantly degrade undermining robust robotic navigation.
Researchers have recently proposed LiDAR–inertial–visual odometry (LIVO) frameworks, that integrate LiDAR, IMU, and camera sensors, to address these challenges. 
This paper extends the FAST-LIVO2-based framework by introducing a hybrid approach that integrates direct photometric methods with descriptor-based feature matching. For the descriptor-based feature matching, this work proposes pairs of ORB with the Hamming distance, SuperPoint with SuperGlue, SuperPoint with LightGlue, and XFeat with the mutual nearest neighbor. The proposed configurations are benchmarked by accuracy, computational cost, and feature tracking stability, enabling a quantitative comparison of the adaptability and applicability of visual descriptors.
The experimental results reveal that the proposed hybrid approach outperforms the conventional sparse-direct method. Although the sparse-direct method often fails to converge in regions where photometric inconsistency arises due to illumination changes, the proposed approach still maintains robust performance under the same conditions. Furthermore, the hybrid approach with learning-based descriptors enables robust and reliable visual state estimation across challenging environments.

\end{abstract}


\titlepgskip=-15pt

\maketitle

\section{Introduction}
\label{sec:introduction}
\PARstart{}{} 
Recently, simultaneous localization and mapping (SLAM) technology has increasingly been employed in autonomous robotic exploration, such as search-and-rescue operations in disaster environments.
The LiDAR–inertial–visual odometry (LIVO) framework, which fuses light detection and ranging (LiDAR), inertial measurement unit (IMU), and camera, has been investigated to achieve robustness and stability in such challenging environments.
For example, FAST-LIVO2~\cite{fastlivo2} is a well-known sparse-direct (SD) LIVO framework that tightly integrates LiDAR, IMU, and camera measurements using an error-state iterated Kalman filter (ESIKF). This framework has demonstrated accurate and robust localization across diverse indoor and outdoor environments, while achieving real-time performance and high-precision mapping.
However, under harsh real-world conditions, conventional LIVO systems often suffer visual degeneration due to reduced visibility caused by heavy smoke and dust, and from high parallax induced by abrupt terrain changes~\cite{motionblur}. In addition, the accuracy of LiDAR measurements can be degraded by collapsed structures or complex obstacles~\cite{genzicp, xicp}. Therefore, visual and LiDAR residuals associated with IMU-propagated states in LIVO systems become distorted, leading to divergence and error accumulation during the ESIKF update process. Such compounded degeneration conditions expose the limitations of direct visual odometry methods widely used in LIVO frameworks.

Direct visual odometry methods~\cite{dso, lsd} rely on the photometric consistency assumption, which assumes that the same 3D point maintains constant brightness across multiple frames. Hence, photometric calibration, compensating for device-specific photometric properties such as camera sensor response, lens vignetting, and image signal processing, is critical to ensure stable tracking performance. However, in real-world environments, illumination changes, such as the sunset, switching lights on and off, or moving shadows, cannot be fully corrected using calibration alone, often significantly degrading performance in direct visual odometry~\cite{direct-cons1, direct-cons2}. Therefore, the visual odometry community has increasingly explored feature-based approaches as an alternative to overcome the limitations of direct methods.

For more robust feature extraction and tracking, feature-based visual odometry methods~\cite{msckf, okvis, vinsmono, openvins} must achieve high-accuracy pose estimation and reliable environmental mapping by applying stable visual information under sufficient lighting conditions. However, in low-light environments, the available visual cues decrease drastically, making feature detection and tracking nearly impossible and often causing the entire system to fail~\cite{feat2}.

As an approach to strengthen feature-based methods, learning-based visual odometry methods~\cite{superpoint, superglue, lightglue, learning1, learning2, learning3, learning4, roft-vins} significantly enhance the robustness and discriminativeness of local descriptors, enabling reliable matching even under challenging conditions, including illumination changes or low-texture environments. Nevertheless, robust keypoint detection across the entire image remains challenging in texture-sparse regions, and reliable matching cannot be ensured under extreme viewpoint changes~\cite{loftr}. Furthermore, Lenc et al.~\cite{lenc2018large} demonstrated that, in tasks where viewpoint invariance is critical, traditional handcrafted methods can still outperform learning-based approaches, revealing that relying solely on learning-based visual odometry remains insufficient to ensure consistent robustness across diverse conditions.

Existing LIVO systems have primarily adopted early, traditional direct methods, as tightly coupled LiDAR–camera integration has recently attracted significant attention~\cite{r3live, r3live++, lviofusion, fastlivo2}. This preference also stems from the perception that photometric error–based alignment more effectively achieves precise sensor fusion between LiDAR and cameras. However, in environments with unstable illumination (e.g., low-light, overexposure, illumination changes, and high parallax) photometric error–based optimization may fail to converge or accumulate substantial drift.
To address these limitations, this study explores the feasibility of incorporating visual feature–based methods and introduces a hybrid approach that combines visual feature–based patch filtering with existing photometric error–based state updates. To evaluate robustness under extreme illumination deficiencies, this work develops a unified LIVO framework that enables a consistent comparison and analysis of diverse visual descriptors under identical datasets and scenarios, comprehensively validating their applicability and effectiveness.

Deep learning–based visual descriptors offer high matching accuracy, but their substantial computational cost has traditionally limited their application in real-time robotic systems. Recent graphics processing unit (GPU)-accelerated embedded platforms, including the NVIDIA Jetson series, have improved on-device inference. However, the practical feasibility of deploying each descriptor still critically depends on its computational load.
For practical applications with diverse real-time requirements, this study quantifies the runtime of several feature extractor–matcher combinations in a unified LIVO framework and analyzes the resulting accuracy–efficiency trade-offs. The results reveal that lightweight combinations, such as XFeat with mutual nearest search (MNN) operate close to real time, whereas heavier deep models such as SuperPoint (SP) with SuperGlue(SG) introduce substantially higher latency.
This evaluation clarifies the computational requirements of learning-based visual modules in embedded settings and offers a practical basis for selecting feature configurations suitable for real-time or resource-constrained LIVO deployments.

This work employs the process of FAST-LIVO2~\cite{fastlivo2} as the baseline to investigate proposed strategies for enhancing a visual module under challenging illumination conditions. The principle contributions of this work are as follows:

\begin{itemize}
\item Through the experiments, this work analyzes the causes of convergence failure and error accumulation in the sparse-direct based ESIKF optimization of FAST-LIVO2 when operating under challenging conditions such as illumination changes and high parallax.

\item This work proposes a hybrid approach by integrating four feature extractor–matcher pairs---ORB~\cite{orb} with the Hamming distance (HD), SP~\cite{superpoint} with SG~\cite{superglue}, SP~\cite{superpoint} with LightGlue (LG)~\cite{lightglue}, and XFeat~\cite{xfeat} with MNN search ---into the sparse-direct based LIVO process.

\item This work systematically compares and analyzes the conventional sparse-direct method and the proposed hybrid approach in terms of accuracy, computational cost, and feature tracking stability, using datasets that include various illumination-deficient scenarios such as low-light, overexposure, illumination changes, and high parallax.
\end{itemize}

\section{Related works}

This section categorizes visual odometry estimation methods applicable to the LIVO framework into three perspectives : direct, feature-based, and learning-based, and reviews the principles of each approach and their performance limitations in odometry estimation.

\subsection{\textbf{Direct Visual Odometry}}

DSO~\cite{dso} is a direct method of visual SLAM that exploits pixel-level photometric information to overcome the instability of feature-based approaches in low-light or low-texture environments. This method employs a sliding-window formulation to optimize camera poses, intrinsic parameters, inverse depths, and photometric correction parameters jointly, achieving high tracking accuracy and robustness. However, the validity range of the photometric consistency assumption is highly limited, making DSO vulnerable to motion blur and heavily dependent on initialization. In addition, as the solar-glare intensity increases, inverse-depth-based tracking tends to fail, highlighting dynamic illumination changes and posing a significant weakness in direct-based SLAM~\cite{direct-cons2}.

Fast semi-direct monocular visual odometry (SVO)~\cite{svo} is a representative approach that combines the strengths of direct and indirect methods to achieve efficient and robust pose estimation. By applying Bayesian filter-based depth estimation and performing bundle adjustment over a limited set of keyframes, SVO maintains computational efficiency while achieving high accuracy and real-time performance. The visual measurement module of FAST-LIVO2 in this study is based on this sparse-direct method.

\subsection{\textbf{Feature-based Visual Odometry}}

The representative feature extractor ORB~\cite{orb} estimates the position and orientation of keypoints using oriented FAST. The orientation information is applied to the rotated BRIEF descriptor to achieve rotation invariance. In addition, by incorporating multiscale detection based on an image pyramid, ORB provides scale invariance while satisfying the requirements of real-time SLAM systems with lightweight and fast computation. However, even such feature-based approaches reveal several limitations in complex real-world environments.

\begin{figure*}[t]
\centering
\includegraphics[width=\textwidth]{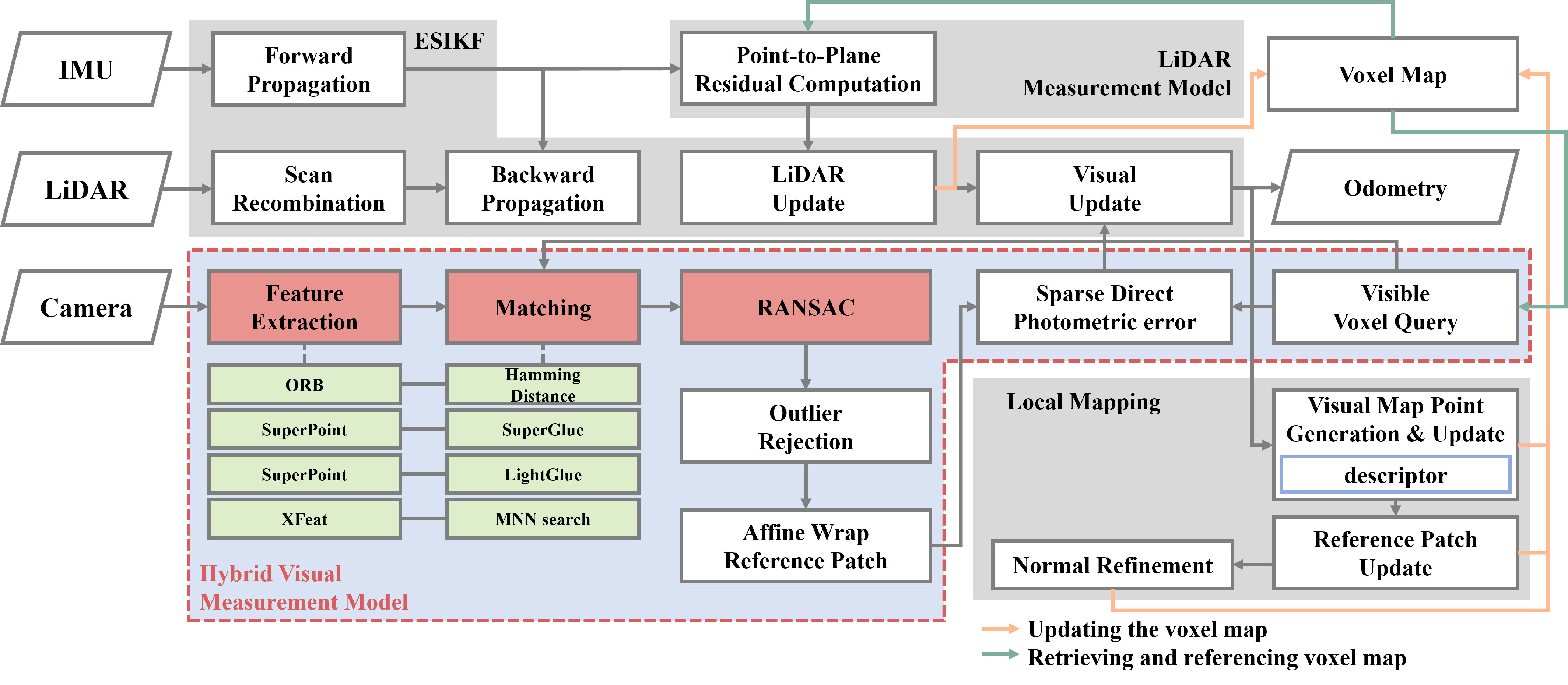}
\caption{Framework for the LiDAR-inertial-visual odometry (LIVO) system built on FAST-LIVO2 to integrate four pairs of visual feature extractors and matchers}
\label{fig1:livo_framework}
\end{figure*}


\begin{figure}[t]
\centering
\includegraphics[width=\linewidth]{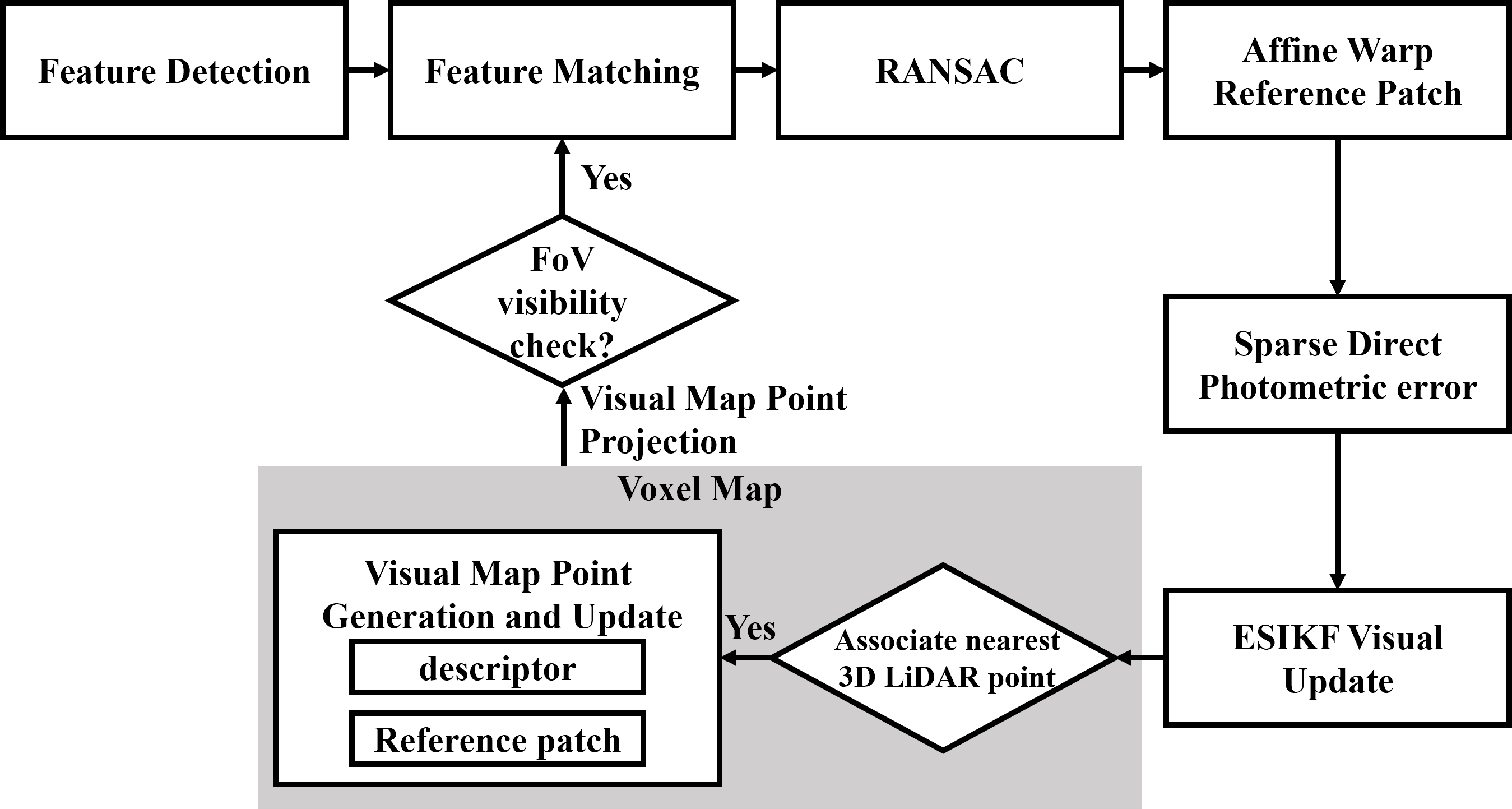}
\caption{Step-by-step hybrid visual measurement update process}
\label{fig:hybrid_visual_update}
\end{figure}

Feature-based methods typically experience a substantial decrease in the number of detectable keypoints under low-light conditions, increasing the risk of matching failures. To mitigate this problem, prior studies have investigated various strategies, including lowering the detection threshold to augment feature counts and applying convolutional neural network–based global contrast enhancement techniques to improve the visibility of dark regions and strengthen feature matching accuracy~\cite{feat2, 10195209}.

\subsection{\textbf{Learning-based Visual odometry}}

SuperPoint~\cite{superpoint} implements a self-supervised framework that jointly learns interest point detection and descriptor extraction. This method employs homographic adaptation, where various homography transformations are applied to input images to generate synthetic training data, substantially improving the repeatability of feature detection and enhancing cross-domain generalization. This approach enables robust feature matching even in unlabeled environments and effectively handles variations in scene, illumination, and viewpoint. 
SuperGlue~\cite{superglue} extends this approach by introducing a deep learning–based local descriptor matching network. By combining self-attention and cross-attention mechanisms, SuperGlue learns both intra- and inter-image context in an end-to-end matching framework. Building on this design, LightGlue~\cite{lightglue} preserves the overall architecture of SuperGlue but introduces a mechanism that predicts the difficulty of each image pair in advance and dynamically adjusts the model size and computational load accordingly. Therefore, LightGlue achieves accuracy comparable to SuperGlue while significantly improving matching speed and efficiency.

XFeat~\cite{xfeat} applies a fast and robust image matching algorithm that preserves as much image resolution as possible while reducing the number of channels in a lightweight convolutional neural network backbone. The model supports sparse and semi-dense matching, where sparse matching is performed using the mutual nearest neighbor search. Compared with transformer-based dense matchers (e.g., LoFTR~\cite{loftr} or AspanFormer~\cite{aspanformer}), XFeat~\cite{xfeat} delivers competitive matching accuracy while being significantly more efficient, making it suitable for real-time SLAM and embedded applications.

Furthermore, hybrid approaches that combine handcrafted techniques with learning-based matching have also been proposed. For instance, in Mix-VIO~\cite{mixvio} extracts keypoints, whereas LightGlue~\cite{lightglue} matches features to enhance robustness against dynamic illumination. In parallel, optical-flow-based tracking is performed for handcrafted features with strong gradients. This design enables Mix-VIO to achieve robust and reliable state estimation even under challenging conditions including rapid motion, which often leads to image blur.

Existing LIVO frameworks~\cite{fastlivo2, r3live, r3live++, dvloam, sdvloam, lviofusion} tend to rely on direct methods, and no prior work has consistently applied and evaluated diverse visual features in the same LIVO framework. To enable a fair quantitative comparison, this work selects ORB~\cite{orb} as a representative handcrafted method, SuperPoint~\cite{superpoint} as a learning-based method, and XFeat~\cite{xfeat} as a recent resource-efficient learning-based method. These feature extractors are combined with widely used matchers (i.e. Hamming distance, SuperGlue~\cite{superglue} and LightGlue~\cite{lightglue}) to form four configurations : ORB + HD, SP + SG, SP + LG, and XFeat + MNN. Using visually challenging datasets that include illumination changes, this work comprehensively evaluate optimization convergence, computational efficiency, and feature-specific performance.

\section{HYBRID SPARSE-DIRECT AND FEATURE-BASED VISUAL ODOMETRY}

Fig.~\ref{fig1:livo_framework} illustrates the overall architecture of the proposed visually-improved LIVO system. The proposed method is built on the FAST-LIVO2 framework and comprises four components: ESIKF, local mapping, the LiDAR measurement model, and the visual measurement model. The proposed visual measurement model is designed to incorporate the sparse-direct method employed in FAST-LIVO2 and integrate feature-based methods. Reliable patches are first selected using a visual feature extractor–matcher pair, after which the sparse-direct method is performed by applying affine warping and minimizing the photometric error.

\subsection{\textbf{Visual-Inertial Odometry OF FAST-LIVO2}}

\subsubsection{Visual Measurement Model}

The proposed method constructs a visual submap from the voxel map in each camera frame, serving as the input to the visual measurement update. One visual map point is selected for every $30\times30$ pixel grid cell, and the set of these points constitutes the visual submap. Each reference patch is warped onto the current frame using an affine transformation, and tracking is performed by minimizing the photometric error.

\subsubsection{Local Mapping}
LiDAR points visible from the current frame are projected onto the image plane to generate and update visual map points. Candidate points are selected if the surrounding image patches display sufficiently large gray-level gradients. For each candidate, the visual state is updated to refine its 3D coordinates, which are re-projected onto the camera image.

\subsection{\textbf{Proposed Hybrid Visual-Inertial Odometry framework}}

\subsubsection{Coarse-to-Fine Visual Measurement Module}

The step-by-step hybrid visual measurement update process is illustrated in Fig.~\ref{fig:hybrid_visual_update}. 
In the hybrid approach, the conventional sparse-direct method is extended by incorporating a feature extractor–matcher module into the visual measurement model. First, visual map points in the camera field of view are retrieved from the voxel map. Next, each 2D–3D correspondence undergoes geometric validation using a RANSAC-based reprojection error under the predicted camera pose, and only inliers are retained. Finally, for inlier points, reference patches are warped onto the current frame using affine transformations, and photometric errors are computed to update the visual state.

\subsubsection{Descriptor-based Local Mapping}

In this study, the visual measurement model exploits only inlier correspondences that have undergone geometric validation. For each 2D inlier keypoint, this study searches in a small $3\times 3$-pixel window centered at the keypoint and select the closest projected LiDAR point in the image space. This patch-based search acts as an implicit distance threshold and handles small reprojection misalignments. If no LiDAR return is available in the $3\times 3$ window (typically due to occlusion or sparsity) the keypoint is discarded for the visual update.

\begin{table}[t]
  \centering
  \caption{Factors in degeneracy of Newer College (NC)~\cite{newercollege}, SubT-MRS (STM)~\cite{superloc}, and MARS-LVIG (ML) datasets}
  \label{tab:dataset_details}
  \begin{tabular}{@{} c c c c @{}}
    \toprule
    \textbf{Dataset} & \textbf{Sequence} & \textbf{Visual Degeneration} & \textbf{LiDAR Degeneration} \\
    \midrule
    \multirow{2}{*}{NC}
        & Quad-Hard   & Planar surfaces      & Well-constrained    \\
        & Maths-Hard  & Planar surfaces      & Well-constrained    \\
    \midrule
    \multirow{2}{*}{STM}
        & Cave1       & Illumination change  & Partial-constrained \\
        & Cave2       & Illumination change  & Partial-constrained \\
    \midrule
    \multirow{2}{*}{ML}
        & AMvalley02  & Low parallax         & Vertical structure loss \\
        & AMvalley03  & Low parallax         & Vertical structure loss \\
    \bottomrule
  \end{tabular}
\end{table}

\section{Experiments}

\subsection{\textbf{Datasets}}

The experiments evaluate the proposed method on the NewerCollege~\cite{newercollege}, SubT-MRS~\cite{superloc}, and MARS-LVIG~\cite{marslvig} datasets.
These datasets cover a broad range of sensing modalities and environmental conditions, 
from multicamera visual–inertial setups, fisheye-based systems, and aerial LiDAR–camera platforms to outdoor campus scenes, low-light subterranean tunnels, feature-sparse corridors, and fast aerial trajectories.
Together, these variations allow the assessment of accuracy improvements and the generalizability of the proposed hybrid visual module across sensor configurations and previously unseen environments.
To compare the performance between the conventional sparse-direct method and the proposed method incorporating patch filtering with visual feature extractor–matcher pairs, this work presents a quantitative analysis in the same LIVO process. Table~\ref{tab:dataset_details} summarizes the LiDAR and visual degeneracy factors of the three datasets.

The Newer College~\cite{newercollege} dataset was collected at the University of Oxford’s New College campus through pedestrian traversal using a various mobile mapping sensors. Among its sequences, Quad-Hard involves aggressive motions, arising from fast walking speed, occasional walking along walls, and instances of overexposure. Maths-Hard also presents challenges for SLAM systems, because it combines rapid walking with abrupt rotations and shaking over largely textureless surfaces. 

The SubT-MRS~\cite{superloc} dataset encompasses challenging environments with diverse structural characteristics, including caves, long corridors, staircases, and tunnels, and was collected with LiDAR, IMU, and fisheye cameras that were synchronized with a precision of 3 ms. 
The cave sequence significantly hinders the performance of visual odometry due to poor illumination. LiDAR odometry also faces significant challenges in this environment because of insufficient geometric features and the repetitiveness, and monotony of the structures. 

The MARS-LVIG~\cite{marslvig} dataset was collected using a quadrotor across diverse environments, including an aero-model airfield, island, rural town, and valley. Each sequence was obtained at different cruising speeds while following the same set of waypoints at a consistent altitude. The HKAirport and HKIsland scenarios were captured in the evening to provide variations in lighting conditions. The AMValley sequence contains visual and LiDAR degeneracy, primarily due to ground-facing flight

\begin{table}[ht]
\centering
\caption{Visual feature extractor–matcher pairs integrated into the LiDAR–inertial–visual odometry framework. \\
}
\label{tab:frontend_variants}
\begin{tabular}{ccc}
\toprule
\textbf{Method} & \textbf{Feature Type} & \textbf{Feature + Matcher} \\
\midrule
SD-only (baseline) & None & None (photometric only)\\
hybrid SD+ORB+HD & Handcrafted & ORB~\cite{orb}+HD\\
hybrid SD+SP+SG & Learning-based & SP~\cite{superpoint}+SG~\cite{superglue}\\
hybrid SD+SP+LG & Learning-based  & SP~\cite{superpoint}+LG~\cite{lightglue}\\
hybrid SD+XFeat+MNN & Learning-based  & XFeat~\cite{xfeat}+MNN\\
\bottomrule
\label{tab:compared_algo}
\end{tabular}
\vspace{-0.3cm}
\begin{tablenotes}
    \item[1] SD: sparse-direct~\cite{svo}, HD: hamming distance, SP: SuperPoint~\cite{superpoint}, SG: SuperGlue~\cite{superglue}, LG: LightGlue~\cite{lightglue} and MNN: mutual nearest neighbor search
\end{tablenotes}    
\vspace{-0.15cm}
\end{table}

\begin{table*}[ht]
\centering
\caption{Main parameters used in the experiments.
For sparse-direct~\cite{svo}, ORB~\cite{orb}, SuperPoint~\cite{superpoint}, SuperGlue~\cite{superglue} and LightGlue~\cite{lightglue}, the parameter settings primarily follow the recommended presets provided by the original authors, with minimal adjustments to fit the LIVO experimental environment. All parameters were applied consistently to ensure a fair comparison.}
\label{tab:fastlivo2_params}

\begin{tabular}{@{}lll p{6cm}@{}}
\toprule
\textbf{Algorithm} & \textbf{Parameter} & \textbf{Value} & \textbf{Description} \\
\midrule

\multirow{3}{*}{Sparse-Direct}
& Pyramid levels & 4 & Number of image pyramid levels \\
& Patch size & 8\,px & Patch size for photometric error calculation \\
& Photometric outlier threshold & 1000 & Threshold for outlier rejection based on the squared photometric error \\
\midrule

\multirow{6}{*}{ORB}
& Max features & 1000 & Maximum number of features per frame \\
& Pyramid levels & 8 & Number of pyramid levels \\
& Scale factor & 1.2 & Scaling ratio between pyramid levels \\
& Border exclusion & 31\,px & Border margin excluded from detection \\
& FAST threshold & 20 & Corner detection sensitivity \\
& Descriptor size & 31\,px & Patch size for descriptor computation \\
\midrule

\multirow{2}{*}{SuperPoint}
& Border size & 4\,px & Border margin excluded from detection \\
& Keypoint confidence threshold & 0.015 (default) & Minimum detector confidence for keypoint acceptance \\
\midrule

\multirow{1}{*}{XFeat}
& Keypoint confidence threshold (default) & 0.001 & Minimum detector confidence for keypoint acceptance \\
\midrule

\multirow{2}{*}{SuperGlue/LightGlue}
& Max matching distance & 4\,px & Maximum pixel distance allowed for matching \\
& Matching confidence threshold & 0.10 (default) & Minimum matcher confidence to accept a correspondence \\

\bottomrule
\end{tabular}
\end{table*}

\subsection{\textbf{Algorithm adopted}}

\setcounter{table}{3}
\begin{table*}[!t]
  \centering
  \caption{Comparison of absolute pose error in translation [m] across datasets for different visual feature extractor–matcher pairs, evaluated after full SE(3) alignment.}
  \label{tab:errors}
  \begin{tabular}{@{} l
      *{4}{cc}   
      *{4}{cc}   
      *{4}{cc}   
    @{}}
    \toprule
    & \multicolumn{4}{c}{Newer College}
    & \multicolumn{4}{c}{Super-Loc}
    & \multicolumn{4}{c}{\makecell[c]{MARS-LVIG\\(w sliding window)}} \\
    \cmidrule(lr){2-5}\cmidrule(lr){6-9}\cmidrule(lr){10-13}
    \makecell[l]{Visual Module\\Configuration}
      & \multicolumn{2}{c}{Quad-Hard}
      & \multicolumn{2}{c}{Maths-Hard}
      & \multicolumn{2}{c}{Cave1}
      & \multicolumn{2}{c}{Cave2}
      & \multicolumn{2}{c}{AMvalley02}
      & \multicolumn{2}{c}{AMvalley03} \\
    \cmidrule(lr){2-3}\cmidrule(lr){4-5}
    \cmidrule(lr){6-7}\cmidrule(lr){8-9}
    \cmidrule(lr){10-11}\cmidrule(lr){12-13}
    & RMSE & Max & RMSE & Max 
    & RMSE & Max & RMSE & Max 
    & RMSE & Max & RMSE & Max  \\
    \midrule
    SD-only (Baseline)       & 0.587 & 1.072 & 0.555 & 1.049 & 0.255 & 1.321 & 12.128 & 21.403 & 1.834 & 12.721 & 8.283 & 13.351 \\
    hybrid SD+ORB+HD        & 0.594 & 0.986 & 0.556 & 1.063 & 0.221 & 0.398 & 11.641 & 20.764 & 1.831 & 9.193 & 5.214 & 15.620 \\
    hybrid SD+SP+SG     & 0.590 & 0.974 & 0.557 & 1.078 & 0.216 & 0.398 & 6.206 & 11.154 & 1.854 & 11.952 & 7.327 & 11.824 \\
    hybrid SD+SP+LG     & 0.591 & 0.975 & 0.557 & 1.037 & {0.221} & 0.418 & 4.292 & 8.194 & 1.779 & {6.913} & 5.892 & \textbf{11.093} \\
    hybrid SD+XFeat+MNN     & \textbf{0.052} & \textbf{0.148} & \textbf{0.061} & \textbf{0.121} & \textbf{0.185} & \textbf{0.344} & \textbf{0.488} & \textbf{1.263} & \textbf{1.105} & \textbf{2.494} & {2.641} & 24.450 \\

    \bottomrule
  \end{tabular}
  \vspace{0.1cm}
  \begin{tablenotes}
        \item[1] SD: sparse-direct~\cite{svo}, HD: hamming distance, SP: SuperPoint~\cite{superpoint}, SG: SuperGlue~\cite{superglue}, LG: LightGlue~\cite{lightglue} and MNN: mutual nearest neighbor search
  \end{tablenotes}    
  \vspace{-0.15cm}
        
\end{table*}

Table~\ref{tab:compared_algo} presents the visual feature extractor–matcher pairs evaluated in this paper. Based on the sparse-direct method, FAST-LIVO2 served as the baseline framework.
The experiments consider handcrafted and deep learning–based feature matching approaches. As a representative handcrafted method, ORB combined with the HD was evaluated. For learning-based approaches, this work integrates SP with SG and SP with LG, and lightweight variants such as XFeat with MNN search. All pairs were incorporated into the same LIVO framework, enabling a consistent and quantitative comparison of their characteristics and performance.

To ensure a fair comparison, this work does not perform dataset-specific parameter tuning.
All feature extractors, matchers, and visual odometry processes were configured using the default or officially recommended settings from their original implementations, and these parameters were kept constant across all datasets.
The sparse-direct (SD) baseline was also evaluated strictly using the original FAST-LIVO2 default presets.
Table~\ref{tab:fastlivo2_params} presents the complete configurations for all evaluated algorithms.
We additionally verified that moderate variations around these default parameters do not lead to meaningful changes in trajectory accuracy.

\section{Result Analysis}
\subsection{\textbf{Implementation and System Configurations}}

\begin{figure*}[htbp]
  \centering

  \begin{minipage}[c]{0.327\textwidth}
    \centering
    \includegraphics[width=\linewidth]{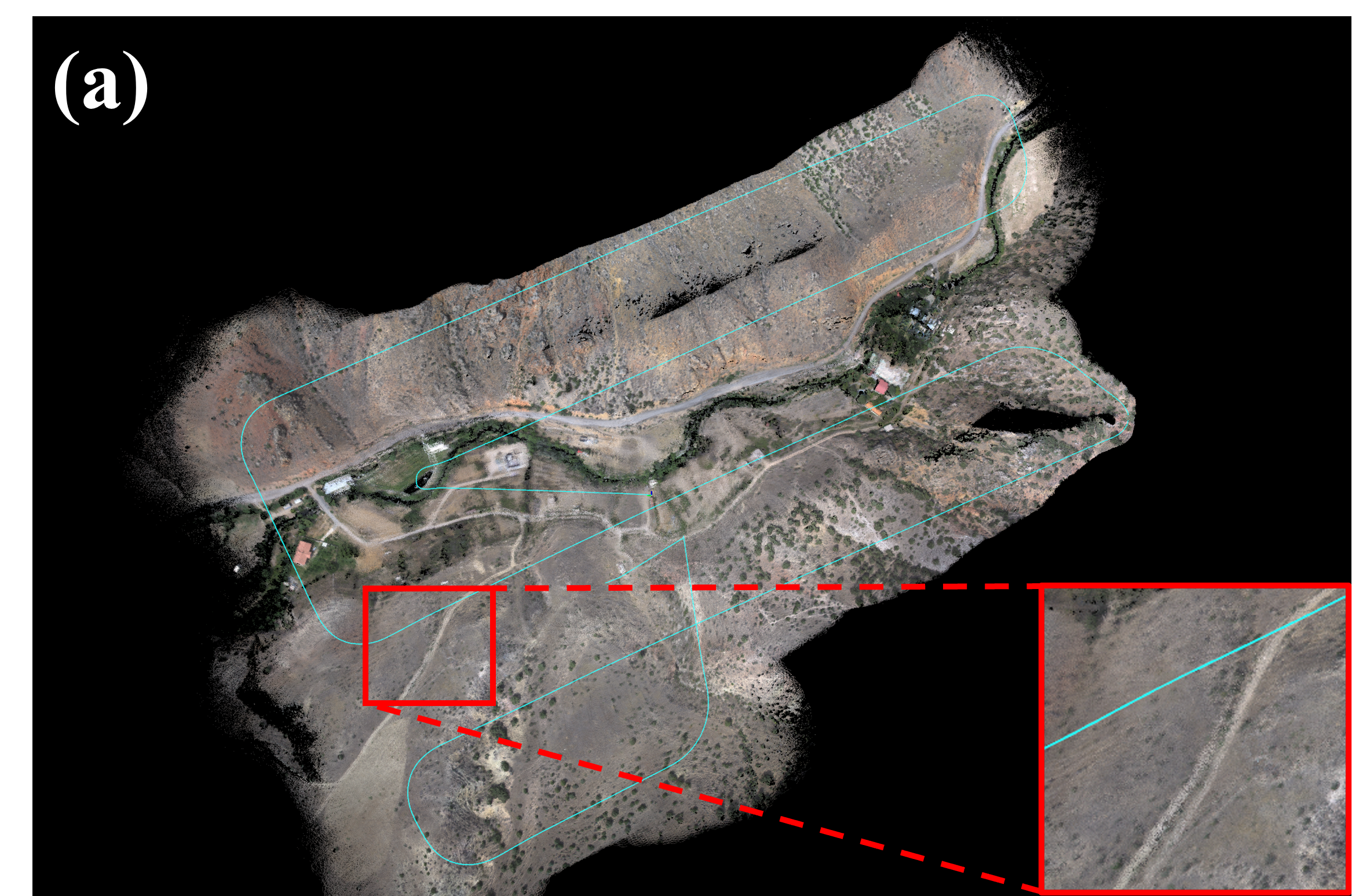}
  \end{minipage}%
  \hspace{-0.008\textwidth}
  \begin{minipage}[c]{0.654\textwidth}
    \centering

    \begin{minipage}[t]{0.495\linewidth}
      \centering
      \includegraphics[width=\linewidth]{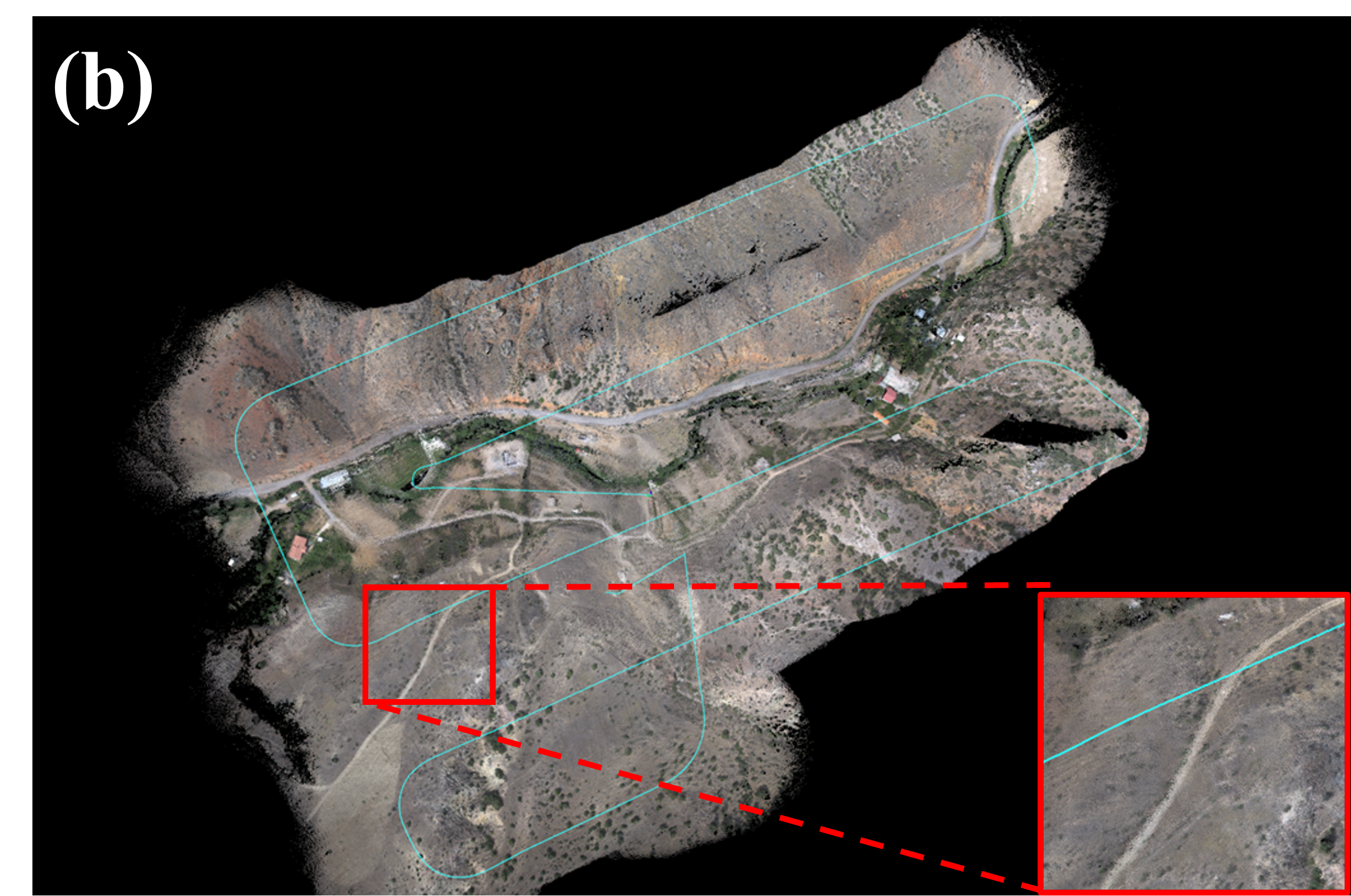}
    \end{minipage}%
    \hspace{-0.01\linewidth}
    \begin{minipage}[t]{0.495\linewidth}
      \centering
      \includegraphics[width=\linewidth]{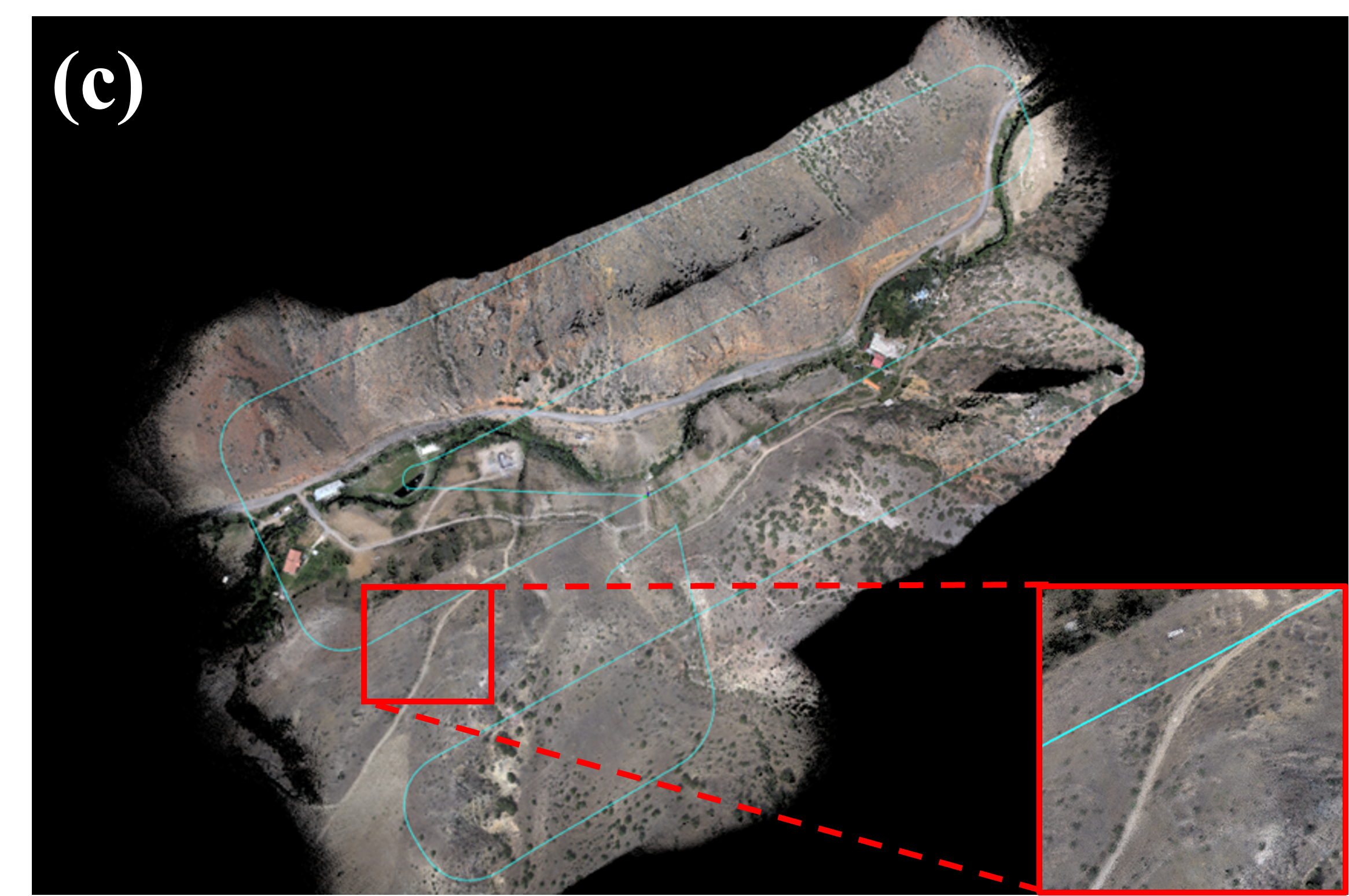}
    \end{minipage}

    \vspace{0.6em}

    \begin{minipage}[t]{0.495\linewidth}
      \centering
      \includegraphics[width=\linewidth]{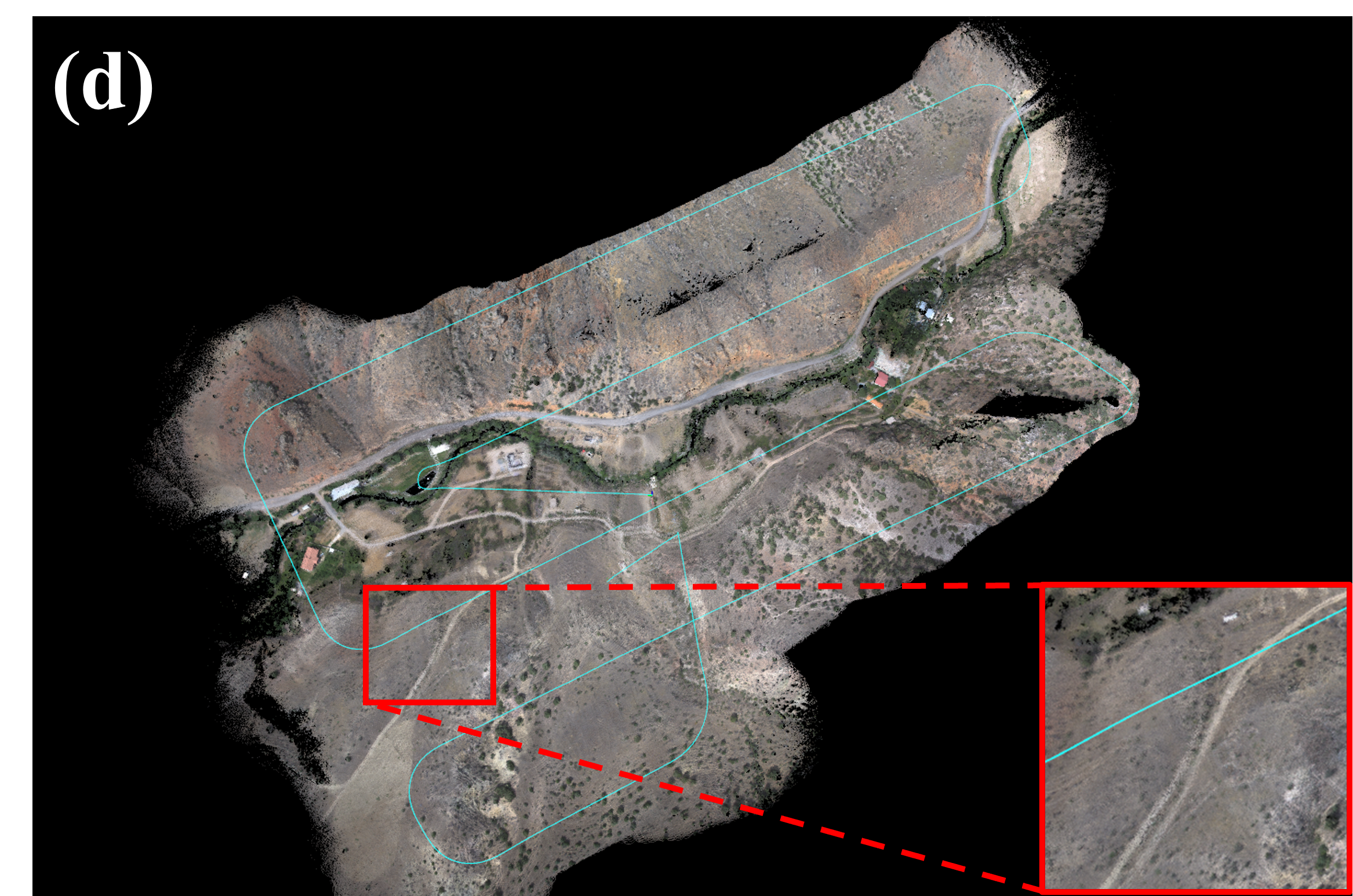}
    \end{minipage}%
    \hspace{-0.01\linewidth}
    \begin{minipage}[t]{0.495\linewidth}
      \centering
      \includegraphics[width=\linewidth]{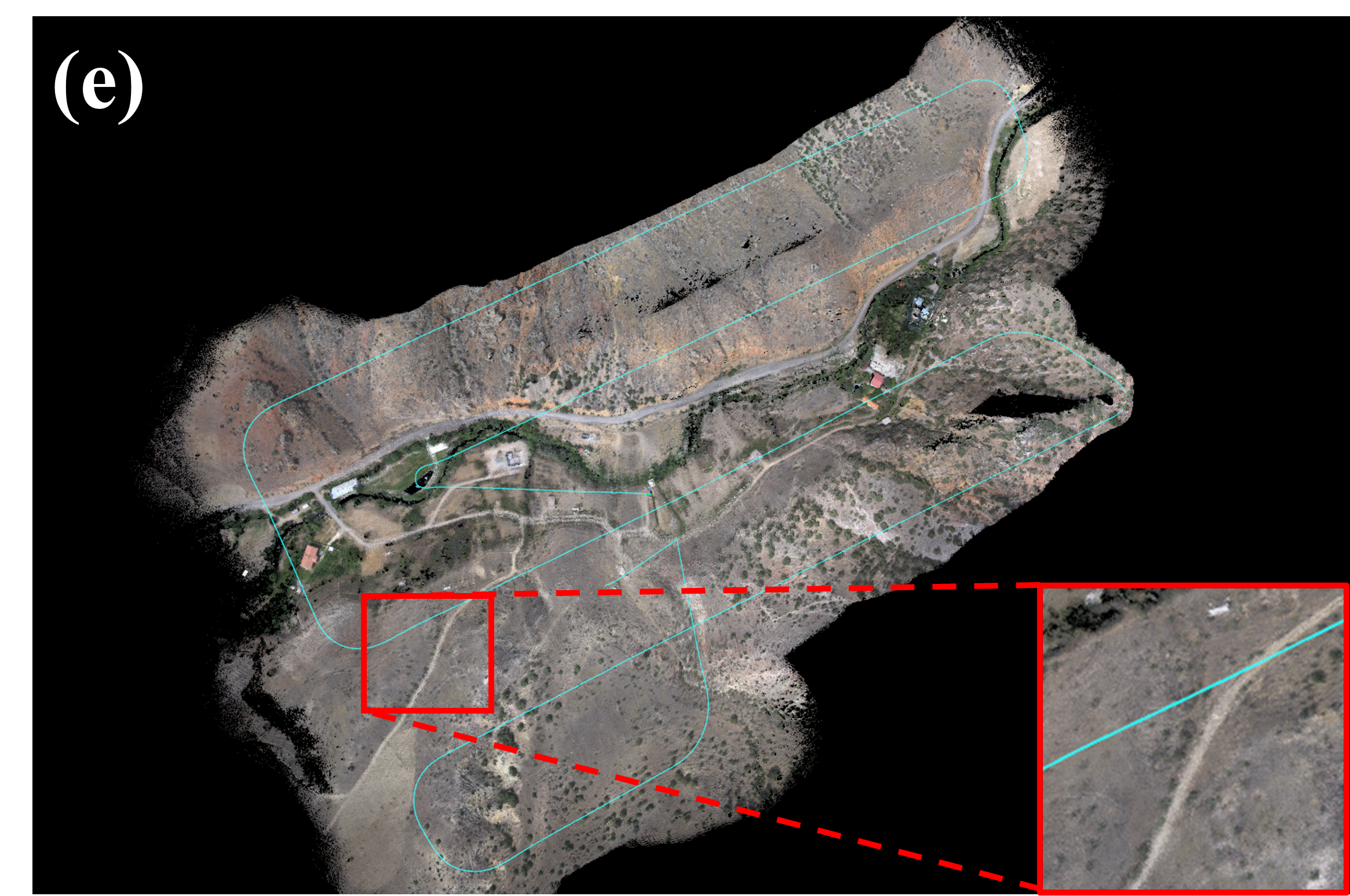}
    \end{minipage}

  \end{minipage}

  \caption{Mapping results on the AMValley03 sequence using sliding-window local mapping: 
  (a) sparse-direct only, 
  (b) ORB~\cite{orb} + Hamming distance, 
  (c) SuperPoint~\cite{superpoint} + SuperGlue~\cite{superglue}, 
  (d) SuperPoint~\cite{superpoint} + LightGlue~\cite{lightglue}, and
  (e) XFeat~\cite{xfeat} + mutual nearest neighbor search. 
  Square red insets reveal zoomed-in point-cloud details, and the blue line traces the UAV’s flight path.}
  \label{fig:visualize}
\end{figure*}

\begin{figure}[t]
\centering
\includegraphics[width=\linewidth]{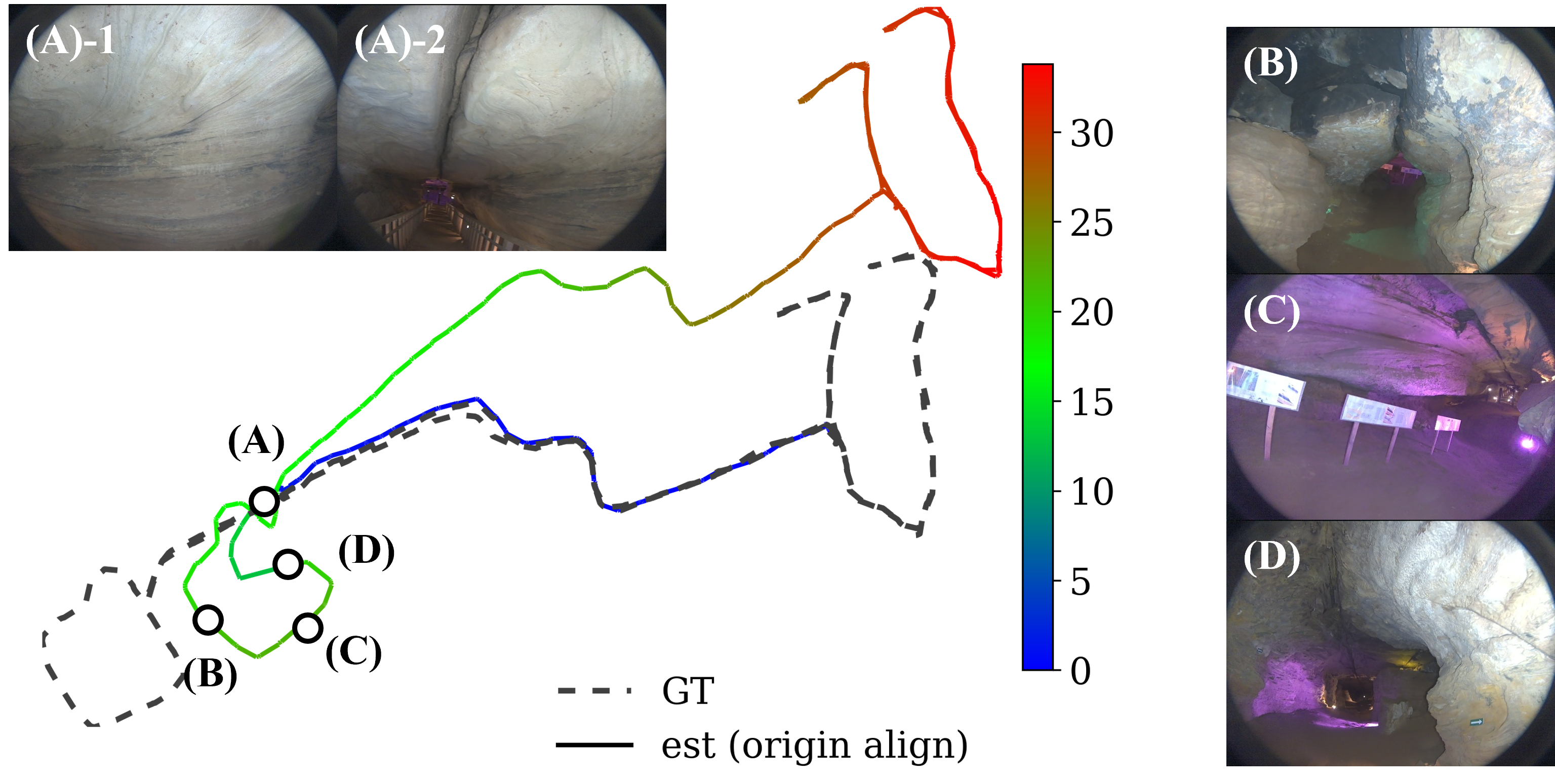}
\caption{Trajectories on the Cave2 sequence. The sparse-direct (SD)-only method exhibits pronounced deviations at specific segments, which are caused by high parallax in (A) and illumination changes in (B)–(D). The color bar indicates the absolute trajectory error (m), and trajectories are aligned by translating the initial poses (origin alignment) to emphasize where and how drift accumulates without global alignment.}
\label{fig1:traj-cave2}
\end{figure}

This work converts the inference code of the SP and XFeat feature extractors, and the SG and LG matchers, from Python to C++ and deploys them using ONNX Runtime 1.16.3 with the CUDA execution provider in FP32 precision. All learning-based models use officially released pretrained weights without fine-tuning, and input images are resized to a fixed resolution of 640×480.
All experiments were conducted on a desktop PC equipped with an Intel Core i9-10900 CPU (10 cores @ 2.8–5.2 GHz), 64 GB RAM, and an NVIDIA GeForce RTX 4090 GPU with 24 GB VRAM.

The MARS-LVIG dataset involves long-duration flights and large-area environments, thus, this work adopts a sliding-window strategy to address the memory constraints. The local map radius was limited to 100 m, and the map was updated every 8 m along the trajectory.

\subsection{\textbf{ACCURACY COMPARISON}}

The Table~\ref{tab:errors} presents the root mean square error (RMSE) of the absolute pose error with respect to translation. The proposed hybrid approach performed the best overall across the three datasets compared with the sparse-direct only method. 

In the Newer College dataset, the hybrid SD+SP+SG and SD+SP+LG variants exhibited slightly lower accuracy than the SD-only method in the Maths-Hard and Quad-Hard sequences. These sequences include rapid rotational motion that induces strong non-linear streak blur and abrupt HDR illumination transitions—conditions that exceed the range of synthetic augmentations used during SuperPoint training. Thus, the descriptor consistency and keypoint repeatability of SuperPoint degrade, leading to minor increases in error.
Nevertheless, the proposed hybrid framework employs feature-based patch filtering, preventing such descriptor instabilities from heavily influencing the sparse-direct photometric update. Therefore, the performance degradation remains limited to only millimeter–centimeter levels, and the overall odometry stability is maintained.
In contrast, hybrid SD+XFeat+MNN demonstrated significant performance gains over the SD-only method, even in sequences where other hybrid variants offered little to no benefit.

\begin{figure*}[t]
  \centering
  \captionsetup[subfloat]{skip=-2.0pt}
  \includegraphics[width=1.0\linewidth]{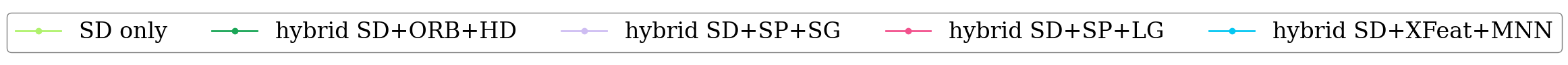}\vspace{-0.6em}

  \begin{minipage}[t]{0.5\linewidth}
    \centering
    \vspace{-0.5em}
    \subfloat{%
      \includegraphics[width=\linewidth]{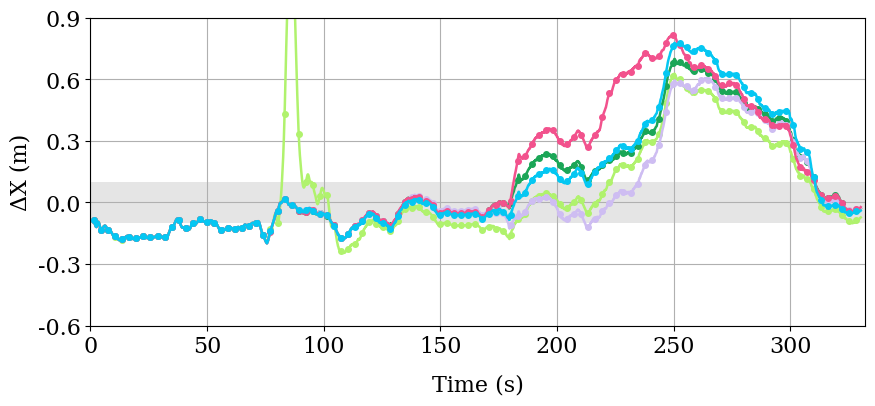}%
      \label{fig:cpu}}
  \end{minipage}%
  \hfill
  \begin{minipage}[t]{0.5\linewidth}
    \centering
    \vspace{-0.5em}
    \subfloat{%
      \includegraphics[width=\linewidth]{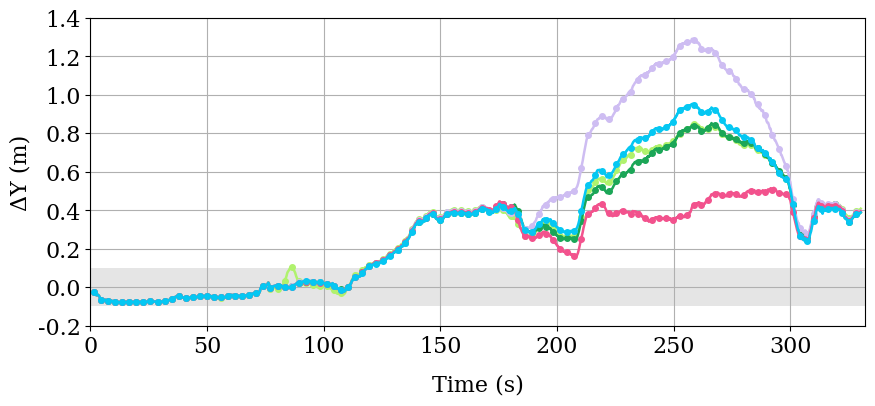}%
      \label{fig:mem}}
  \end{minipage}

  \vspace{-0em}

  \begin{minipage}[t]{0.5\linewidth}
    \centering
    \subfloat{%
      \includegraphics[width=\linewidth]{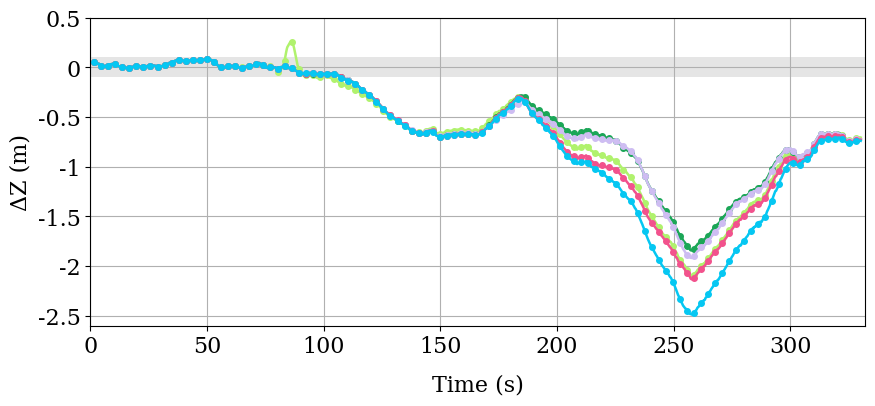}%
      \label{fig:gpu}}
  \end{minipage}%
  \hfill
  \begin{minipage}[t]{0.5\linewidth}
    \centering
    \subfloat{%
      \includegraphics[width=\linewidth]{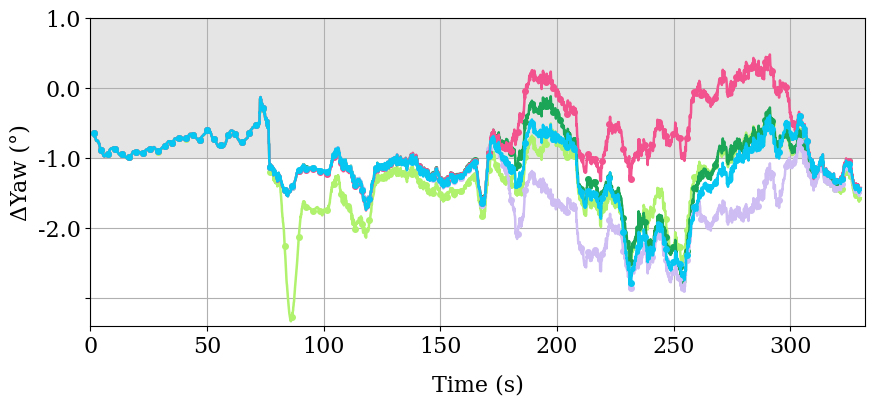}%
      \label{fig:gpu_mem}}
  \end{minipage}
  \caption{
  Resulting trajectories on the Cave1 sequence. The estimated trajectories were aligned with the ground truth using full SE(3) alignment (rotation and translation). Gray segments indicate regions where the position error falls within the range of $-0.1\,\mathrm{m} to 0.1\,\mathrm{m}$, and in the yaw plot, gray bands correspond to yaw errors from $-1^{\circ} to 1^{\circ}$. Here, SD, HD, SP, SG, LG, and MNN refer to Sparse-Direct~\cite{svo}, hamming distance, SuperPoint~\cite{superpoint}, SuperGlue~\cite{superglue}, LightGlue~\cite{lightglue}, and mutual nearest neighbor search, respectively. }

  \label{fig:traj_error}
\end{figure*}


The SubT-MRS dataset exhibits insufficient geometric features, making the advantages of the hybrid approach especially beneficial in this environment.
Among the hybrid approaches, hybrid  SD+XFeat+MNN achieved the lowest errors in the Cave1 and 2 sequences, outperforming hybrid SD+XFeat+MNN, with hybrid SD+SP+LG achieving the next-best performance.
Furthermore, although the SD-only method suffered from pronounced trajectory deviations in the Cave2 sequence due to high parallax and illumination changes, (Fig.~\ref{fig1:traj-cave2}), the handcrafted hybrid SD+ORB+HD maintained odometry convergence with comparatively reduced errors, indicating improved robustness. This improvement can be attributed to the geometric validation stage, which suppresses incorrect visual measurements before photometric optimization.

In the MARS-LVIG dataset, relatively high absolute pose errors were observed overall, partly because the sequences cover wide areas. The AMvalley03 sequence, collected along the same waypoint trajectory as AMvalley02 but at a higher cruising speed, displayed larger errors.

Although AMvalley02 and AMvalley03 are vulnerable to visual degeneration due to ground-facing flight, the hybrid SD+ORB+HD performed relatively robustly on AMvalley03, whereas the learning-based hybrids SD+SP+SG and SD+SP+LG yielded comparatively lower accuracy. This result can be attributed to the large-scale environment, enabling the extraction of a sufficient number of visual features.

In contrast, the learning-based hybrids SD+SP+SG and SD+SP+LG, revealed notably lower performance. Two factors may account for this. First, deep learning–based feature extractors are often limited by detection confidence threshold settings, which can restrict the number of detected points and reduce the available inliers after geometric verification. Second, in segments with monotonous viewing directions, the lack of structural diversity in feature matches may limit the effectiveness of learning-based matchers. Lenc et al.~\cite{lenc2018large} reported that variations in the number of detections caused by threshold settings can directly affect repeatability, and they observed that learning-based detectors are typically more robust to illumination changes but remain less competitive than handcrafted methods under significant viewpoint changes.
Nevertheless, hybrid SD+XFeat+MNN consistently performed best overall. Moreover, in Fig.~\ref{fig:visualize}, hybrid SD+ORB+HD and SD+SP+SG produced the most visually seamless reconstructions in mountain mapping.

Fig.~\ref{fig:traj_error} illustrates the temporal evolution of the position and yaw errors for each method on the Cave1 sequence. Overall, the handcrafted hybrid SD+ORB+HD perform relatively stably across the trajectories, whereas the learning-based hybrid SD+SP+SG and SD+SP+LG display larger deviations, particularly in the lateral and yaw estimates.

During the initial section (about $0\!$ - $\!150\ \mathrm{s}$), the SD-only method displays pronounced yaw oscillations, whereas the other configurations remain relatively stable, although they reveal a systematic offset of about $-1^\circ$.
During the middle phase (about $150\!$ - $\!250\ \mathrm{s}$), the error drift becomes more pronounced, and clear differences emerge across methods and error components. In the lateral directions, $\Delta X$ and yaw errors were largest for hybrid SD+SP+SG, whereas hybrid SD+SP+LG achieved the smallest values. Conversely, for $\Delta Y$ and $\Delta Z$, hybrid SD+SP+LG produces the largest deviations, whereas hybrid SD+SP+SG recorded the lowest result. Overall, hybrid SD+ORB+HD achieved the least vertical error ($\Delta Z$) among the configurations.

In the later phase (about $250\!$ - $\!300\ \mathrm{s}$), the error trends primarily follow the same patterns observed in the middle phase, but the differences between configurations become progressively smaller. By the end of the sequence, all methods converge toward similar error levels across all configurations.

\begin{figure}[!htbp]
  \centering
  \captionsetup[subfloat]{skip=-2.0pt}
  \includegraphics[width=1.0\linewidth]{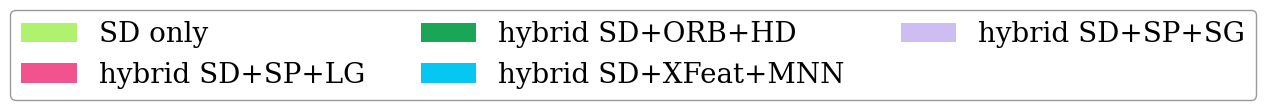}\vspace{-0.6em}


  \begin{minipage}[t]{0.5\linewidth}
    \centering
    \vspace{-0.5em}
    \subfloat[CPU Usage (\%)]{%
      \includegraphics[width=\linewidth]{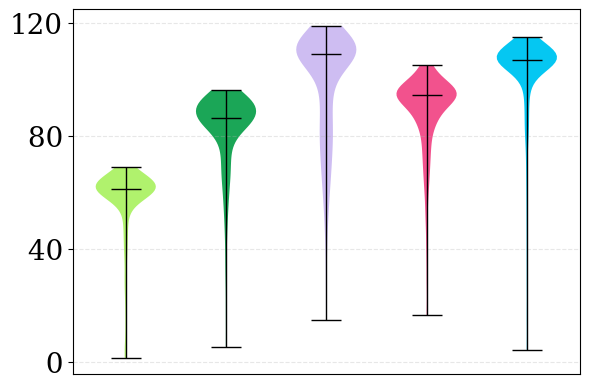}%
      \label{fig:cpu}}
  \end{minipage}%
  \hfill
  \begin{minipage}[t]{0.5\linewidth}
    \centering
    \vspace{-0.5em}
    \subfloat[Memory Usage (\%)]{%
      \includegraphics[width=\linewidth]{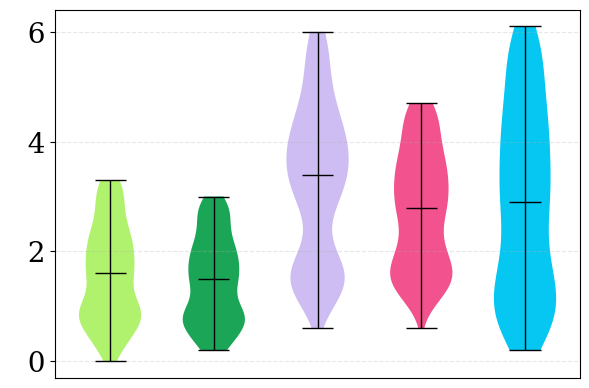}%
      \label{fig:mem}}
  \end{minipage}

  \vspace{-0em} 

  \begin{minipage}[t]{0.5\linewidth}
    \centering
    \subfloat[GPU Usage (\%)]{%
      \includegraphics[width=\linewidth]{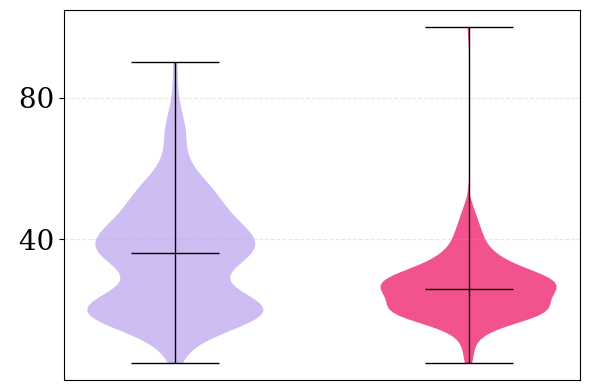}%
      \label{fig:gpu}}
  \end{minipage}%
  \hfill
  \begin{minipage}[t]{0.5\linewidth}
    \centering
    \subfloat[GPU Memory (MiB)]{%
      \includegraphics[width=\linewidth]{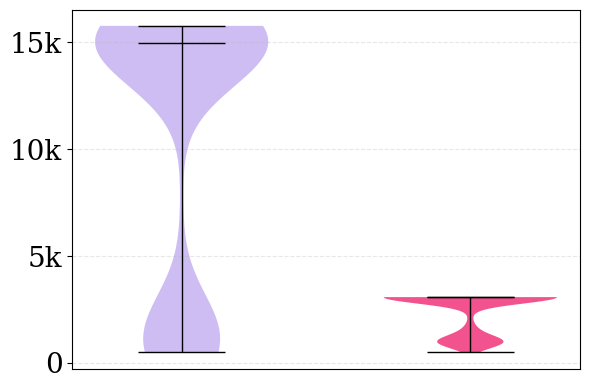}%
      \label{fig:gpu_mem}}
  \end{minipage}

  \caption{Comparison of (a) CPU usage, (b) memory usage, (c) GPU usage, and (d) GPU memory usage on the Cave1 sequence. The SD-only and hybrid SD+ORB+HD are purely CPU-based methods and do not use a GPU (no GPU using or GPU memory consumption). The CPU usage is reported as a percentage of total CPU utilization, where 100\% corresponds to the full usage of a single core. Memory usage indicates the resident set size (RSS) divided by the total physical memory (64\,GB RAM in this experiment). Here, SD, HD, SP, SG, LG, and MNN refer to Sparse-Direct~\cite{svo}, hamming distance, SuperPoint~\cite{superpoint}, SuperGlue~\cite{superglue}, LightGlue~\cite{lightglue}, and mutual nearest neighbor search, respectively.}
  \label{fig:resource_using}
\end{figure}

\subsection{\textbf{COMPUTATIONAL COST COMPARISON}}

\begin{figure*}[t]
  \centering

  \includegraphics[width=0.62\textwidth]{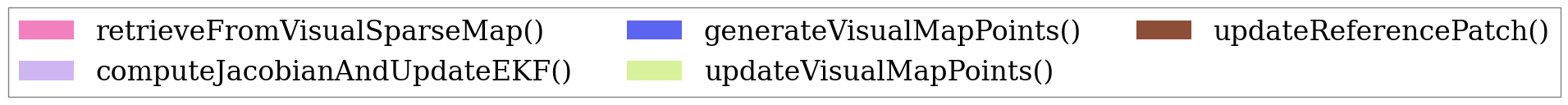}
  \vspace{-0.5em}

  \begin{minipage}[t]{0.5\textwidth}
    \centering
    \subfloat[SD-only]{%
      \includegraphics[width=\linewidth]{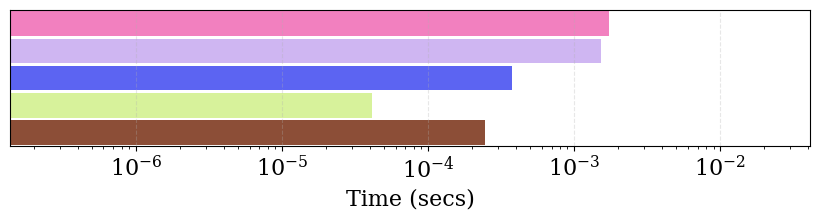}%
      \label{fig:time_direct}}
  \end{minipage}

  \vspace{-0.2em}

  \noindent
  \begin{minipage}[t]{0.5\textwidth}
    \centering
    \subfloat[hybrid SD+ORB+HD]{%
      \includegraphics[width=\linewidth]{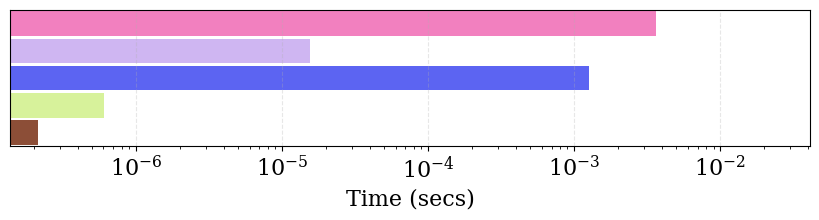}%
      \label{fig:time_orb}}
  \end{minipage}%
  \hfill
  \begin{minipage}[t]{0.5\textwidth}
    \centering
    \subfloat[hybrid SD+SP+SG]{%
      \includegraphics[width=\linewidth]{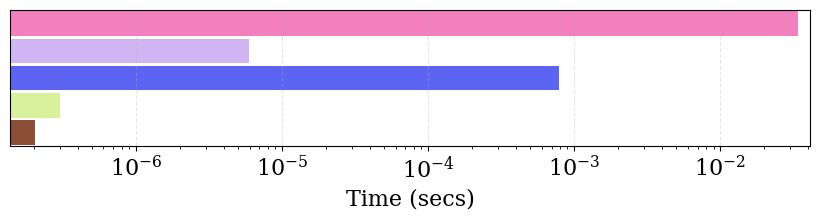}%
      \label{fig:time_sg}}
  \end{minipage}

  \vspace{-0.1em}

  \noindent
  \begin{minipage}[t]{0.5\textwidth}
    \centering
    \subfloat[hybrid SD+SP+LG]{%
      \includegraphics[width=\linewidth]{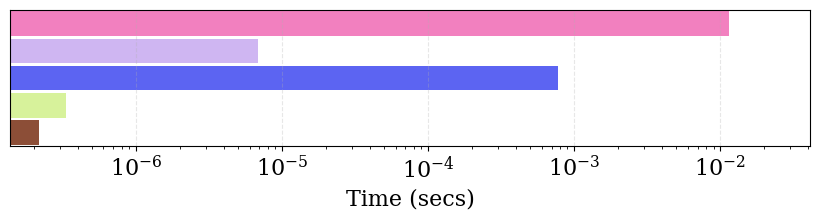}%
      \label{fig:time_lg}}
  \end{minipage}%
  \hfill
  \begin{minipage}[t]{0.5\textwidth}
    \centering
    \subfloat[hybrid SD+XFeat+MNN]{%
      \includegraphics[width=\linewidth]{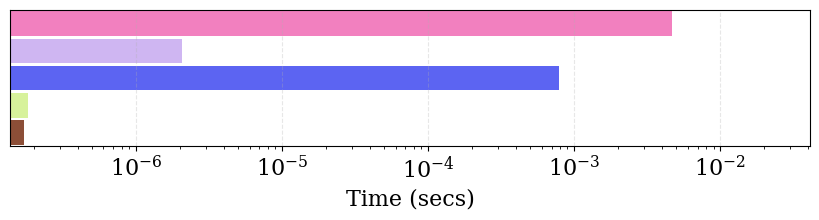}%
      \label{fig:time_mnn}}
  \end{minipage}

  \caption{Frame-wise processing time for each visual feature extractor-matcher configuration on the Cave1 sequence, decomposed into five key stages (summarized in Table.~\ref{tab:vio_stages}). The y-axis represents the average processing time per frame for each stage on a logarithmic scale. The total time shown corresponds to the overall computation time for all VIO processing stages, excluding visualization overhead. Here, SD, HD, SP, SG, LG, and MNN refer to Sparse-Direct~\cite{svo}, hamming distance, SuperPoint~\cite{superpoint}, SuperGlue~\cite{superglue}, LightGlue~\cite{lightglue}, and mutual nearest neighbor search, respectively.}
  \label{fig:processing_time}
\end{figure*}

Fig.~\ref{fig:resource_using} compares the CPU usage, memory usage, GPU usage, and GPU memory consumption across visual module configurations. 
In terms of CPU usage, the SD-only configuration achieved the lowest mean and variance, confirming its lightweight nature because it performs only photometric error–based alignment without feature extraction or matching. In contrast, the hybrid configurations incur a higher CPU load due to the additional operations of feature extraction and matching. Among them, hybrid SD+ORB+HD maintains stable usage around 55\%, whereas the learning-based hybrids impose heavier computational demands. For example, hybrid SD+SP+LG remains close to hybrid SD+ORB+HD in terms of median usage but has sporadic peaks, and hybrid SD+SP+SG reaches peak usage close to 80\% owing to the computationally expensive attention-based architecture of SuperGlue.
Moreover, hybrid SD+XFeat+MNN records slightly higher CPU usage than hybrid SD+ORB+HD due to the descriptor extraction and nearest-neighbor search.

\begin{table}[t]
\centering
\caption{Peak memory consumption of different methods on the Cave1 sequence.
Peak values correspond to the maximum observed memory usage during runtime.
GPU memory is reported only for GPU-enabled methods, including the hybrid SD+SP+SG and SD+SP+LG configurations.
Here, SD, HD, SP, SG, LG, and MNN refer to Sparse-Direct~\cite{svo}, hamming distance, SuperPoint~\cite{superpoint}, SuperGlue~\cite{superglue}, LightGlue~\cite{lightglue}, and mutual nearest neighbor search, respectively.
}
\label{tab:peak_memory}
\begin{tabular}{lcc}
\toprule
\textbf{Method} 
& \makecell{\textbf{Peak CPU Memory}\\\textbf{(\%)}} 
& \makecell{\textbf{Peak GPU Memory}\\\textbf{(MiB)}} \\
\midrule
SD only                     & 3.3 & --    \\
hybrid SD+ORB+HD            & 3.0 & --    \\
hybrid SD+SP+LG             & 4.7 & 3058  \\
hybrid SD+SP+SG             & 6.0 & 15742 \\
hybrid SD+XFeat+MNN         & 6.1 & --    \\
\bottomrule
\end{tabular}
\end{table}

\begin{table}[ht]
\centering
\caption{Stage-wise functions of the visual module execution process. Each function corresponds to a major processing stage, with its role summarized in the right column. The same modular structure is applied in Fig.~\ref{fig:processing_time} to analyze the frame processing time of different feature extractor–matcher pairs.}
\label{tab:vio_stages}
\begin{tabular}{l@{}|l@{}}
\toprule
\textbf{Functions} & \textbf{Description} \\
\midrule

retrieveFromVisualSparseMap()
  & \makecell[l]{Extract candidate visual map points \\ from the current frame} \\
\midrule

computeJacobianAndUpdateEKF()
  & \makecell[l]{Compute Jacobian for photometric error \\ and update EKF state} \\
\midrule

generateVisualMapPoints()
  & \makecell[l]{Generate new visual map points via  \\ voxel raycasting and outlier rejection} \\
\midrule

updateVisualMapPoints()
  & \makecell[l]{Add visual map points to the voxel map} \\
\midrule

updateReferencePatch()
  & \makecell[l]{Update reference patch for selected map \\ points based on patch score} \\

\bottomrule
\end{tabular}
\end{table}

The overall memory usage remains modest across all configurations, distributed within about $0\%$ to $\!6\%$ of the 64~GiB system memory ($\approx0.4$ to $4$~GiB). The SD-only and hybrid SD+ORB+HD methods have the lowest demand, with median usage around $1\%$ to $2\%$, reflecting the lightweight nature of handcrafted approaches. In contrast, the learning-based hybrids SD+SP+SG and SD+SP+LG, increase the memory demand to about $4\%$ on average due to feature extraction and matching. Similarly, hybrid SD+XFeat+MNN also operate in the $3\%$ to $4\%$ range, with the latter demonstrating slightly higher usage than the former, but remaining below 4\%.

\begin{figure*}[t]
  \centering

  \includegraphics[width=0.95\linewidth]{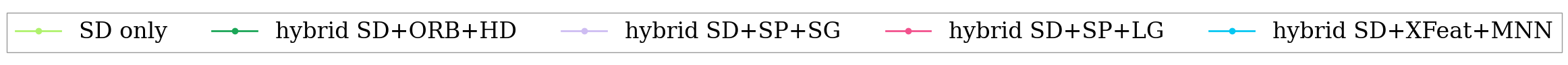}
  \vspace{-1.25em}

  \subfloat[]{%
    \parbox{\textwidth}{\centering
      \textbf{(A) SparseMapSize}\par\vspace{0.3em}
      \includegraphics[width=0.49\linewidth]{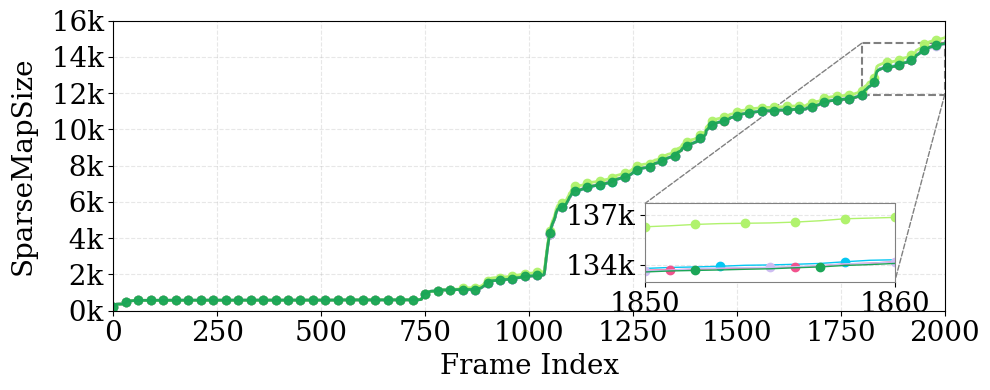}\hspace{-0.015\linewidth}%
      \includegraphics[width=0.49\linewidth]{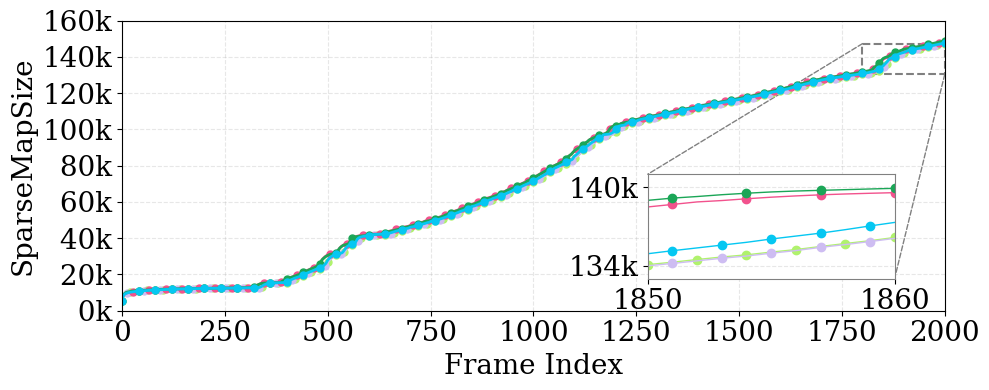}\par
      \vspace{0.2em}
      \footnotesize
      (i) Cave1 sequence of the SubT-MRS dataset~\cite{superloc}\hfill
      (ii) AMValley03 sequence of the MARS-LVIG dataset~\cite{marslvig} (w/o sliding window)
    }%
  }\par
  \vspace{-0.5em}

  \subfloat[]{%
    \parbox{\textwidth}{\centering
      \textbf{(B) FeatureNum}\par\vspace{0.3em}
      \includegraphics[width=0.49\linewidth]{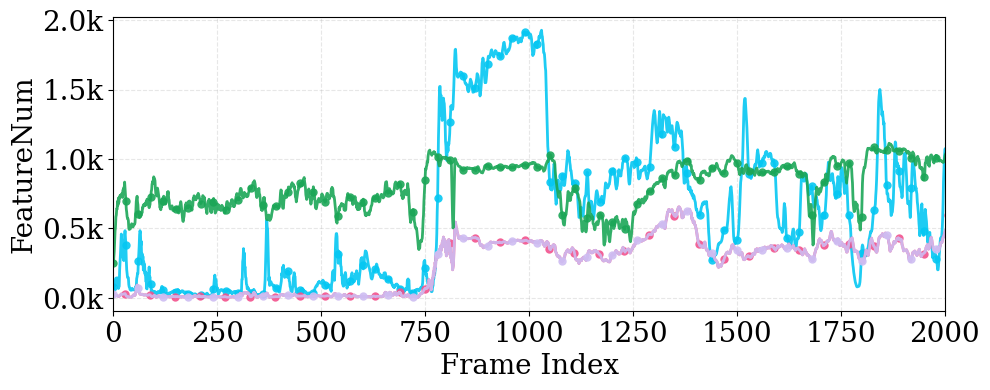}\hspace{-0.015\linewidth}%
      \includegraphics[width=0.49\linewidth]{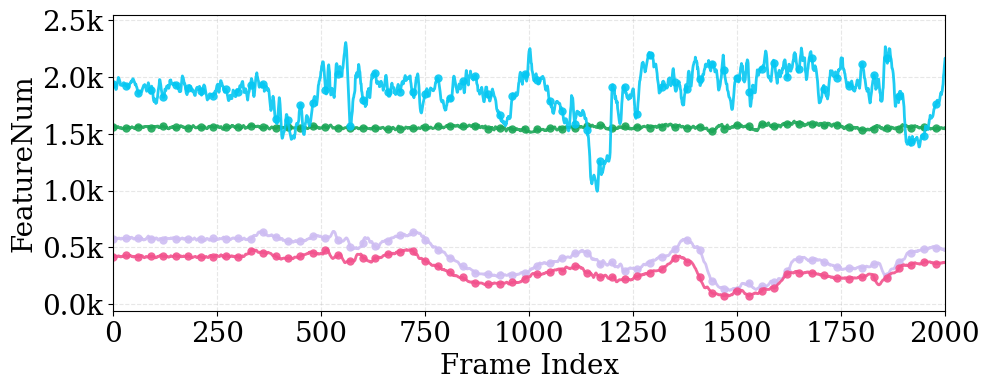}\par
      \vspace{0.2em}
      \footnotesize
      (iii) Cave1 sequence of the SubT-MRS dataset~\cite{superloc}\hfill
      (iv) AMValley03 sequence of the MARS-LVIG dataset~\cite{marslvig} (w/o sliding window)
    }%
  }\par

  \includegraphics[width=0.95\linewidth]{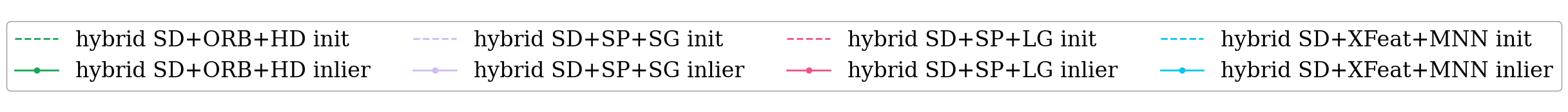}
  \vspace{-1.4em}

  \subfloat[]{%
    \parbox{\textwidth}{\centering
      \textbf{(C) Initial vs.\ Inlier Matches}\par\vspace{0.3em}
      \includegraphics[width=0.49\linewidth]{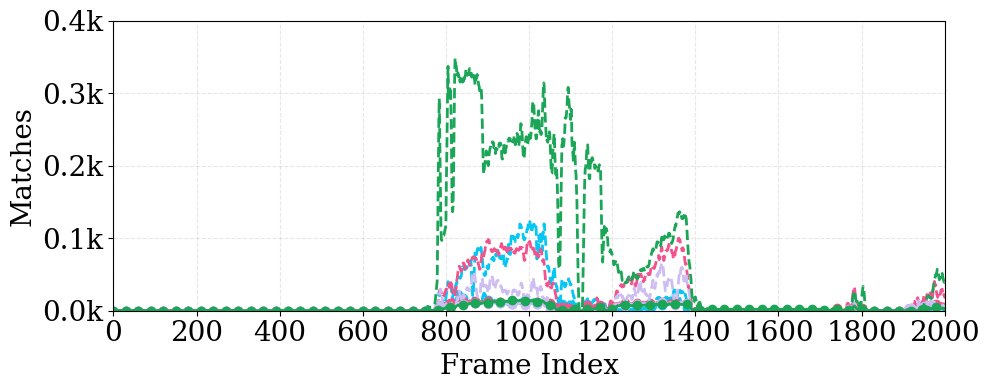}\hspace{-0.015\linewidth}%
      \includegraphics[width=0.49\linewidth]{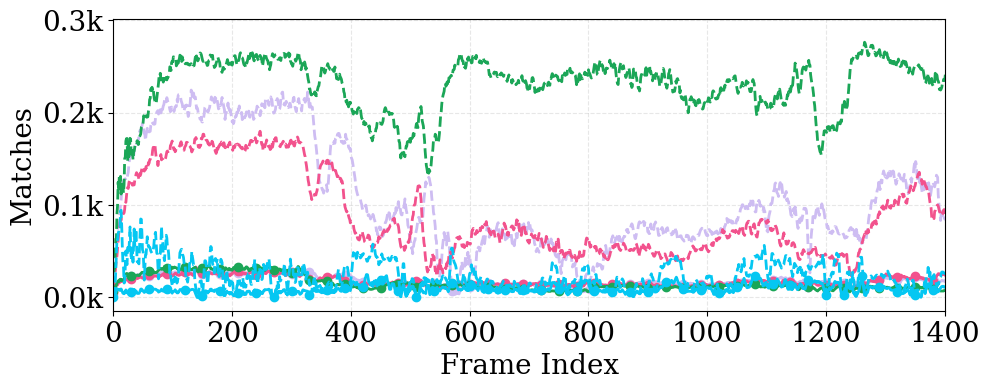}\par
      \vspace{0.2em}
      \footnotesize
      (v) Cave1 sequence of the SubT-MRS dataset~\cite{superloc}\hfill
      (vi) AMValley03 sequence of the MARS-LVIG dataset~\cite{marslvig} (w/o sliding window)
    }%
  }

  \caption{Comparison of mapping efficiency and matching stability for each feature extractor–matcher combination.
  (a) Number of accumulated 3D visual map points (SparseMapSize),
  (b) number of features detected per frame, and
  (c) temporal changes in the number of initial and inlier matches.
  The SD-only configuration does not extract features; hence, plots for (b) and (c) are unavailable.
  Dashed lines indicate the number of initial matches; solid lines with markers represent inlier matches.
  Inliers are obtained via RANSAC-based geometric verification using a 2D–3D reprojection error under the predicted camera pose.
  Here, SD, HD, SP, SG, LG, and MNN refer to Sparse-Direct~\cite{svo}, hamming distance, SuperPoint~\cite{superpoint}, SuperGlue~\cite{superglue}, LightGlue~\cite{lightglue}, and mutual nearest neighbor search, respectively.}
  \label{fig:resource_comparison}
\end{figure*}

In terms of GPU usage, purely CPU-based methods such as the SD-only, hybrid SD+ORB+HD, and hybrid SD+XFeat+MNN methods, do not consume GPU resources. 
In contrast, hybrid SD+SP+SG and hybrid SD+SP+LG rely on GPU acceleration for feature extraction and matching, leading to significantly higher utilization. Among them, hybrid SD+SP+SG exhibits a higher median usage and larger variance, indicating a heavier GPU load, whereas hybrid SD+SP+LG achieved a lower median but a narrower distribution, with peaks reaching $70\%\!$ to $\!80\%$. Overall, LightGlue achieved more efficient GPU utilization than SuperGlue, despite both being learning-based matchers, confirming that LightGlue-based hybrids offer a better trade-off between accuracy and resource efficiency.

In terms of GPU memory usage, the differences are even more pronounced. The hybrid SD+SP+SG method consumes up to about $15$~GiB, primarily because it processes all keypoint pairs exhaustively. In contrast, hybrid SD+SP+LG requires around $2$ to $4$~GiB, $40\%\!$ to $\!50\%$ lower than that for hybrid SD+SP+SG. These results emphasize the efficiency of the LightGlue architecture compared to SuperGlue in hybrid sparse-direct pipelines.

To complement these temporal profiles, Table~\ref{tab:peak_memory} summarizes the peak GPU memory consumption of GPU-enabled methods, defined as the maximum memory usage observed during runtime.

\subsection{\textbf{PROCESSING-TIME COMPARISON}}
Fig.~\ref{fig:processing_time} presents the frame processing time for each visual feature extractor–matcher pair, decomposed across the stages of the visual module. Table~\ref{tab:vio_stages} summarizes the detailed roles of each stage, and the y-axis in Fig.~\ref{fig:processing_time} indicates the average time required to process a single frame at each stage on a logarithmic scale.

Among all configurations, hybrid SD+SP+SG exhibited the highest overall processing time. This overhead is most evident in the stage of generating visual map points, where the self-attention operations of SuperGlue lead to a sharp increase in the computational cost. In contrast, ORB relies on lightweight binary descriptors with simple matching, whereas LightGlue retains the self-attention structure but reduces the computational burden via optimization, such as using keypoint pruning and confidence-based early stopping. 
The hybrid SD+XFeat+MNN method has a processing profile comparable to ORB, because it also relies on lightweight matching, with only minor overhead introduced by descriptor extraction. 
The SD-only method also performs state estimation on all reference patches, but the relatively infrequent depth-based map point generation limits its cost. In general, hybrid methods require more time for map point generation than the SD-only method, because they perform explicit matching between all image features and LiDAR points. In particular, hybrid SD+ORB+HD incurs an even higher cost because each feature also undergoes a geometric verification step following the initial matching. Additionally, learning-based methods generally produce fewer visual map points, resulting in reduced computational burden in the reference patch update stage. Moreover, hybrid methods rely on prevalidated keypoints from the matching stage, maintaining a relatively lightweight Jacobian computation and state update.

\subsection{\textbf{FEATURE TRACKING STABILITY COMPARISON}}

Fig.~\ref{fig:resource_comparison} illustrates how the three metrics measured change as the VIO process updates at 10~Hz. These metrics are the cumulative number of 3D visual map points, number of extracted features, and number of established matches.
As presented in Fig.~\ref{fig:resource_comparison}(b, c), hybrid SD+ORB+HD extracts the largest number of features but suffers from lower and more fluctuating matching counts, indicating reduced overall stability. In contrast, hybrid SD+SP+SG and hybrid SD+SP+LG generate fewer features, yet maintain consistent numbers of matches with higher quality, leading to superior tracking robustness.

The effect of scene characteristics is also evident. In the Cave1 sequence, the repetitive structure and poor illumination limit visual features, resulting in the SD-only method producing more sparse map points than the hybrid methods. Nevertheless, hybrid configurations achieve slightly better odometry accuracy than the sparse-direct baseline, demonstrating their advantages even in such environments.
In contrast, the AMValley03 sequence, captured from a ground-facing viewpoint with abundant visual features, allows the hybrid methods to surpass the SD-only method in terms of the quantity of sparse map points and, more importantly, odometry accuracy. In this setting, hybrid SD+ORB+HD and hybrid SD+SP+SG produced the largest number of visual map points, whereas hybrid SD+XFeat+MNN hold an intermediate level of points, and hybrid SD+SP+LG and the sparse-direct baseline yielded fewer points. However, in terms of accuracy, hybrid SD+XFeat+MNN achieved the best odometry performance. Further, hybrid SD+ORB+HD, hybrid SD+SP+SG, and hybrid SD+SP+LG provided comparable mid-level accuracy. The sparse-direct baseline remained the least accurate. These results indicate that hybrid methods provide consistent improvements in feature-deficient conditions, and their advantages become more pronounced in feature-rich environments, such as AMValley03.

\section{Conclusion}
\setlength{\parskip}{0pt} 
This study empirically demonstrates that the sparse-direct method, commonly adopted in existing LIVO systems, suffers from convergence failure and error accumulation under visually challenging conditions, such as illumination changes and high parallax. 

Building on recent advances in GPU-accelerated embedded platforms that improve the feasibility of on-device deep learning computation, this work presents a hybrid visual module designed to integrate multiple feature extractor–matcher pairs, including ORB + HD, SuperPoint + SuperGlue, SuperPoint + LightGlue, and  XFeat + mutual nearest neighboor search into the sparse-direct process. This work presents a quantitative evaluation across approaches under visual challenging conditions, including illumination changes and high parallax resulting from abrupt maneuvers.

The experimental results revealed that hybrid SD+XFeat+\\MNN consistently achieved the most robust performance across datasets, delivering the highest odometry accuracy and stable matching under diverse conditions. In particular, hybrid SD+XFeat+MNN achieved competitive accuracy without requiring GPU inference, demonstrating its practicality on resource-constrained platforms. Among the other learning-based methods, hybrid SD+SP+SG and hybrid SD+SP+LG provided superior matching quality and more stable state estimation than the direct baseline, provided that at least 2 GiB of GPU memory was available. Notably, hybrid SD+SP+LG reduced GPU memory consumption by $40\%\!$ to $\!50\%$ compared with hybrid SD+SP+SG, while achieving an even lower overall RMSE. In contrast, handcrafted approaches such as hybrid SD+ORB+HD sometimes exhibited greater robustness in scenarios with limited viewpoint changes. This outcome suggests that learning-based features may generalize less effectively when viewpoint variation is limited. 
Across all datasets, compassing diverse sensor configurations and varied operating environments, the proposed hybrid visual module exhibited consistent performance trends, indicating strong generalizability regardless of the sensor and environmental configurations.

The FAST-LIVO2 process runs at 10 Hz, with the VIO update requiring about 3.9 ms per frame. Lightweight descriptors such as XFeat+MNN increase the VIO processing time to 5 to 7 ms, which maintains the overall system at close to 10 Hz. In contrast, heavier matchers such as SuperPoint+SuperGlue and SuperPoint+LightGlue raise the VIO latency to 12 to 35 ms, causing the overall system rate to drop below real-time requirements.

Therefore, to construct a more stable LIVO framework, it is crucial to adopt a hybrid structure in which deep learning–based features are employed for initial patch filtering, followed by sparse-direct state updates. This design helps mitigate local convergence and abnormal optimization behavior arising from the initial pose estimation errors. The study findings are expected to serve as practical guidance for the design of visual modules in LIVO systems, particularly in selecting feature extractor–matcher pairs and devising sensor fusion strategies.~\cite{xfeat}

\bibliographystyle{IEEEtran}
\bibliography{main}

@STRING{ral  = {IEEE Robot. Automat. Lett.} }

@STRING{tro  = {IEEE Trans. Robot.} }

@STRING{ijrr = {Int. J. Robot. Res.} }

@STRING{icra   = {Proc. IEEE Int. Conf. Robot. Automat. (ICRA)} }

@STRING{iccas  = {Proc. Int. Conf. Control, Automat. Syst.} }

@STRING{tpami = {IEEE Trans. Pattern Anal. Mach. Intell.} }

@STRING{cvpr  = {Proc. IEEE/CVF Conf. Comput. Vis. Pattern Recognit. (CVPR)} }

@STRING{cvprw  = {Proc. IEEE/CVF Conf. Comput. Vis. Pattern Recognit. Workshops (CVPRW)} }

@STRING{eccv  = {Proc. Eur. Conf. Comput. Vis.} }

@STRING{accv  ={Proc. Asian Conf. Pattern Recognit.} }

@STRING{iccv  = {Proc. IEEE Int. Conf. Comput. Vis. (ICCV)} }

@STRING{neurips = {Adv. Neural Inf. Process. Syst.} }

@STRING{bmvc = {Proc. Brit. Mach. Vis. Conf.(BMVC)} }

@STRING{itcst = {IEEE Trans. Control Syst. Technol.} }

@ARTICLE{dso,
  author={Engel, Jakob and Koltun, Vladlen and Cremers, Daniel},
  journal=tpami, 
  title={Direct Sparse Odometry}, 
  year={2018},
  volume={40},
  number={3},
  pages={611-625},
  doi={10.1109/TPAMI.2017.2658577}}

@INPROCEEDINGS{lsd,
  author    = {J. Engel and T. Sch{\"o}ps and D. Cremers},
  title     = {{LSD-SLAM: Large-scale direct monocular SLAM}},
  booktitle = eccv,
  address   = {Cham, Switzerland},
  publisher = {Springer},
  year      = {2014},
  pages     = {834--849},
  doi       = {10.1007/978-3-319-10605-2_54}
}

@INPROCEEDINGS{direct-cons1,
  author={Wu, Xiaolong and Pradalier, Cédric},
  booktitle=icra, 
  title={Illumination Robust Monocular Direct Visual Odometry for Outdoor Environment Mapping}, 
  year={2019},
  volume={},
  number={},
  pages={2392-2398},
  doi={10.1109/ICRA.2019.8793607}}

@INPROCEEDINGS{msckf,
  author={Mourikis, Anastasios I. and Roumeliotis, Stergios I.},
  booktitle=icra, 
  title={A Multi-State Constraint {Kalman} Filter for Vision-aided Inertial Navigation}, 
  year={2007},
  volume={},
  number={},
  pages={3565-3572},
  doi={10.1109/ROBOT.2007.364024A}}

@ARTICLE{vinsmono,
  author={Qin, Tong and Li, Peiliang and Shen, Shaojie},
  journal=tro, 
  title={{VINS-Mono}: A Robust and Versatile Monocular Visual-Inertial State Estimator}, 
  year={2018},
  volume={34},
  number={4},
  pages={1004-1020},
  doi={10.1109/TRO.2018.2853729}}

@INPROCEEDINGS{openvins,
  author={Geneva, Patrick and Eckenhoff, Kevin and Lee, Woosik and Yang, Yulin and Huang, Guoquan},
  booktitle=icra, 
  title={{OpenVINS}: A Research Platform for Visual-Inertial Estimation}, 
  year={2020},
  volume={},
  number={},
  pages={4666-4672},
  doi={10.1109/ICRA40945.2020.9196524}}

@inproceedings{okvis,
  title     = {Keyframe-Based Visual–Inertial Odometry Using Nonlinear Optimization},
  author    = {Leutenegger, Simon and Lynen, Simon and Bosse, Michael and Siegwart, Roland and Furgale, Paul},
  booktitle = ijrr,
  volume={34},
  number={3},
  pages     = {314--334},
  year      = {2015},
}

@Article{mixvio,
AUTHOR = {Yuan, Huayu and Han, Ke and Lou, Boyang},
TITLE = {{Mix-VIO}: A Visual Inertial Odometry Based on a Hybrid Tracking Strategy},
JOURNAL = {Sensors},
VOLUME = {24},
YEAR = {2024},
NUMBER = {16},
ARTICLE-NUMBER = {5218},
PubMedID = {39204913},
ISSN = {1424-8220},
DOI = {10.3390/s24165218}
}

@INPROCEEDINGS{superpoint,
  author={DeTone, Daniel and Malisiewicz, Tomasz and Rabinovich, Andrew},
  booktitle=cvprw, 
  title={{SuperPoint}: Self-Supervised Interest Point Detection and Description}, 
  year={2018},
  month={Jun.}, 
  volume={},
  number={},
  pages={224-236},
  doi={10.1109/CVPRW.2018.00060}}

@InProceedings{lightglue,
    author    = {Lindenberger, Philipp and Sarlin, Paul-Edouard and Pollefeys, Marc},
    title     = {{LightGlue}: Local Feature Matching at Light Speed},
    booktitle = iccv,
    month     = {Oct.},
    year      = {2023},
    pages     = {17627-17638}
}

@INPROCEEDINGS{orb,
  author={Rublee, Ethan and Rabaud, Vincent and Konolige, Kurt and Bradski, Gary},
  booktitle=iccv, 
  title={{ORB}: An efficient alternative to {SIFT} or {SURF}}, 
  year={2011},
  month={Nov.},
  volume={},
  number={},
  pages={2564-2571},
  doi={10.1109/ICCV.2011.6126544}}

@INPROCEEDINGS{superglue,
  author={Sarlin, Paul-Edouard and DeTone, Daniel and Malisiewicz, Tomasz and Rabinovich, Andrew},
  booktitle=cvpr, 
  title={{SuperGlue}: Learning Feature Matching With Graph Neural Networks}, 
  year={2020},
  month={Jun.},
  volume={},
  number={},
  pages={4937-4946},
  doi={10.1109/CVPR42600.2020.00499}}

@INPROCEEDINGS{aspanformer,
  author    = {H. Chen and Z. Luo and L. Zhou and Y. Tian and M. Zhen and T. Fang and D. McKinnon and Y. Tsin and L. Quan},
  title     = {{ASpanFormer}: Detector-Free Image Matching with Adaptive Span Transformer},
  booktitle = eccv,
  address   = {Cham, Switzerland},
  publisher = {Springer},
  year      = {2022},
  pages     = {20--36},
  doi       = {10.1007/978-3-031-19824-3_2}
}

@ARTICLE{fastlivo2,
  author={Zheng, Chunran and Xu, Wei and Zou, Zuhao and Hua, Tong and Yuan, Chongjian and He, Dongjiao and Zhou, Bingyang and Liu, Zheng and Lin, Jiarong and Zhu, Fangcheng and Ren, Yunfan and Wang, Rong and Meng, Fanle and Zhang, Fu},
  journal=tro, 
  title={{FAST-LIVO2}: Fast, Direct {LiDAR}–Inertial–Visual Odometry}, 
  year={2025},
  volume={41},
  number={},
  pages={326-346},
  doi={10.1109/TRO.2024.3502198}}

@InProceedings{loftr,
    author    = {Sun, Jiaming and Shen, Zehong and Wang, Yuang and Bao, Hujun and Zhou, Xiaowei},
    title     = {{LoFTR}: Detector-Free Local Feature Matching With Transformers},
    booktitle = cvpr,
    month     = {Jun.},
    year      = {2021},
    pages     = {8922-8931}
}

@INPROCEEDINGS{svo,
  author={Forster, Christian and Pizzoli, Matia and Scaramuzza, Davide},
  booktitle=icra, 
  title={{SVO}: Fast semi-direct monocular visual odometry}, 
  year={2014},
  volume={},
  number={},
  pages={15-22},
  doi={10.1109/ICRA.2014.6906584}}

@InProceedings{superloc,
    author    = {Zhao, Shibo and Gao, Yuanjun and Wu, Tianhao and Singh, Damanpreet and Jiang, Rushan and Sun, Haoxiang and Sarawata, Mansi and Qiu, Yuheng and Whittaker, Warren and Higgins, Ian and Du, Yi and Su, Shaoshu and Xu, Can and Keller, John and Karhade, Jay and Nogueira, Lucas and Saha, Sourojit and Zhang, Ji and Wang, Wenshan and Wang, Chen and Scherer, Sebastian},
    title     = {{SubT-MRS Dataset}: Pushing {SLAM} Towards All-weather Environments},
    booktitle = cvpr,
    month     = {Jun.},
    year      = {2024},
    pages     = {22647-22657}
}

@misc{newercollege,
      title={Multi-Camera {LiDAR} Inertial Extension to the Newer College Dataset}, 
      author={Lintong Zhang and Marco Camurri and David Wisth and Maurice Fallon},
      journal={arXiv preprint arXiv:2112.08854},
      year={2022},
      month={May.},
      eprint={2112.08854},
      archivePrefix={arXiv},
      primaryClass={cs.RO},
      url={https://arxiv.org/abs/2112.08854}, 
}

@article{marslvig,
  author  = {Hao Li and others},
  title   = {{MARS-LVIG} dataset: A multi-sensor aerial robots {SLAM} dataset for {LiDAR}-visual-inertial-{GNSS} fusion},
  journal = ijrr,
  year    = {2024},
  volume  = {43},
  number  = {8},
  pages   = {1114--1127},
  doi     = {10.1177/02783649241227968}
}

@INPROCEEDINGS{r3live++,
  author={Lin, Jiarong and Zhang, Fu},
  booktitle=tpami, 
  title={{R$^{3}$LIVE++}: A Robust, Real-Time, Radiance Reconstruction Package With a Tightly-Coupled {LiDAR}-Inertial-Visual State Estimator}, 
  year={2024},
  volume={46},
  number={12},
  pages={11168-11185},
  doi={10.1109/TPAMI.2024.3456473}}

@INPROCEEDINGS{r3live,
  author={Lin, Jiarong and Zhang, Fu},
  booktitle=icra, 
  title={{R$^{3}$LIVE}: A Robust, Real-time, {RGB}-colored, {LiDAR}-Inertial-Visual tightly-coupled state Estimation and mapping package}, 
  year={2022},
  volume={},
  number={},
  pages={10672-10678},
  doi={10.1109/ICRA46639.2022.9811935}}

@Article{dvloam,
AUTHOR = {Wang, Wei and Liu, Jun and Wang, Chenjie and Luo, Bin and Zhang, Cheng},
TITLE = {{DV-LOAM}: Direct Visual {LiDAR} Odometry and Mapping},
JOURNAL = {Remote Sensing},
VOLUME = {13},
YEAR = {2021},
}

@ARTICLE{sdvloam,
  author={Yuan, Zikang and Wang, Qingjie and Cheng, Ken and Hao, Tianyu and Yang, Xin},
  journal=tpami, 
  title={{SDV-LOAM}: Semi-Direct Visual–{LiDAR} Odometry and Mapping}, 
  year={2023},
  volume={45},
  number={9},
  pages={11203-11220},
  doi={10.1109/TPAMI.2023.3262817}}

@ARTICLE{lviofusion,
  author={Zhang, Hongkai and Du, Liang and Bao, Sheng and Yuan, Jianjun and Ma, Shugen},
  journal=ral, 
  title={{LVIO-Fusion}:Tightly-Coupled {LiDAR}-Visual-Inertial Odometry and Mapping in Degenerate Environments}, 
  year={2024},
  volume={9},
  number={4},
  pages={3783-3790},
  doi={10.1109/LRA.2024.3371383}}

@inproceedings{direct-cons2,
  author    = {Ruben Gomez-Ojeda and Zichao Zhang and Javier Gonzalez-Jiménez and Davide Scaramuzza},
  title     = {Learning-based Image Enhancement for Visual Odometry in Challenging {HDR} Environments},
  booktitle = icra,
  year      = {2018},
  pages     = {805--811},
  isbn      = {978-1-5386-3081-5},
}

@article{genzicp,
   title={{GenZ-ICP}: Generalizable and Degeneracy-Robust {LiDAR} Odometry Using an Adaptive Weighting},
   volume={10},
   ISSN={2377-3774},
   DOI={10.1109/lra.2024.3498779},
   number={1},
   journal=ral,
   publisher={Institute of Electrical and Electronics Engineers (IEEE)},
   author={Lee, Daehan and Lim, Hyungtae and Han, Soohee},
   year={2025},
   month=jan, pages={152–159} }

@ARTICLE{xicp,
  author={Tuna, Turcan and Nubert, Julian and Nava, Yoshua and Khattak, Shehryar and Hutter, Marco},
  journal=tro, 
  title={{X-ICP}: Localizability-Aware {LiDAR} Registration for Robust Localization in Extreme Environments}, 
  year={2024},
  volume={40},
  number={},
  pages={452-471},
  doi={10.1109/TRO.2023.3335691}}

@InProceedings{motionblur,
    author    = {Liu, Peidong and Zuo, Xingxing and Larsson, Viktor and Pollefeys, Marc},
    title     = {{MBA-VO}: Motion Blur Aware Visual Odometry},
    booktitle = iccv,
    month     = {Oct.},
    year      = {2021},
    pages     = {5550-5559}
}

@InProceedings{learning1,
author = {Dusmanu, Mihai and Rocco, Ignacio and Pajdla, Tomas and Pollefeys, Marc and Sivic, Josef and Torii, Akihiko and Sattler, Torsten},
title = {{D2-Net}: A Trainable {CNN} for Joint Description and Detection of Local Features},
booktitle = cvpr,
month = {Jun.},
year = {2019}
}

@InProceedings{learning2,
author = {Luo, Zixin and Zhou, Lei and Bai, Xuyang and Chen, Hongkai and Zhang, Jiahui and Yao, Yao and Li, Shiwei and Fang, Tian and Quan, Long},
title = {{ASLFeat}: Learning Local Features of Accurate Shape and Localization},
booktitle = cvpr,
month = {Jun.},
year = {2020},
pages = {6588--6597}
}

@inproceedings{learning3,
  author    = {Jerome Revaud and Philippe Weinzaepfel and C{\'{e}}sar Roberto de Souza and
               Martin Humenberger},
  title     = {{R2D2:} Repeatable and Reliable Detector and Descriptor},
  booktitle = neurips,
  year      = {2019},
  pages     = {12}
}

@InProceedings{learning4,
    author    = {Tian, Yurun and Balntas, Vassileios and Ng, Tony and Barroso-Laguna, Axel and Demiris, Yiannis and Mikolajczyk, Krystian},
    title     = {{D2D}: Keypoint Extraction with Describe to Detect Approach},
    booktitle = accv,
    month     = {Nov.},
    year      = {2020},
    pages     = {9308-9326}
}

@INPROCEEDINGS{feat2,
  author={Shyam, Pranjay and Bangunharcana, Antyanta and Kim, Kyung-Soo},
  booktitle=iccas, 
  title={Retaining Image Feature Matching Performance Under Low Light Conditions}, 
  volume={},
  number={},
  pages={1079-1085},
  doi={10.23919/ICCAS50221.2020.9268426}}

@inproceedings{lenc2018large,
  author    = {Karel Lenc and Andrea Vedaldi},
  title     = {Large scale evaluation of local image feature detectors on homography datasets},
  booktitle = bmvc,
  year      = {2018},
  pages={1-13}
}

@InProceedings{xfeat,
    author    = {Potje, Guilherme and Cadar, Felipe and Araujo, Andr\'e and Martins, Renato and Nascimento, Erickson R.},
    title     = {XFeat: Accelerated Features for Lightweight Image Matching},
    booktitle = cvpr,
    month     = {June},
    year      = {2024},
    pages     = {2682-2691}
}

@INPROCEEDINGS{roft-vins,
  author={Park, Sanghyun and Han, Soohee},
  booktitle=iccas, 
  title={{ROFT-VINS}: Robust Feature Tracking-based Visual-Inertial State Estimation for Harsh Environment}, 
  year={2024},
  volume={},
  number={},
  pages={508-513},
  doi={10.23919/ICCAS63016.2024.10773196}}

@INPROCEEDINGS{10195209,
  author={Lee, Hoyong and Lee, Hakjun and Kwak, Inveom and Sung, Chiwon and Han, Soohee},
  journal=itcst, 
  title={Effective Feature-Based Downward-Facing Monocular Visual Odometry}, 
  year={2024},
  volume={32},
  number={1},
  pages={266-273},
  doi={10.1109/TCST.2023.3294843}}

\begin{IEEEbiography}[{\includegraphics[width=1in,height=1.25in,clip,keepaspectratio]{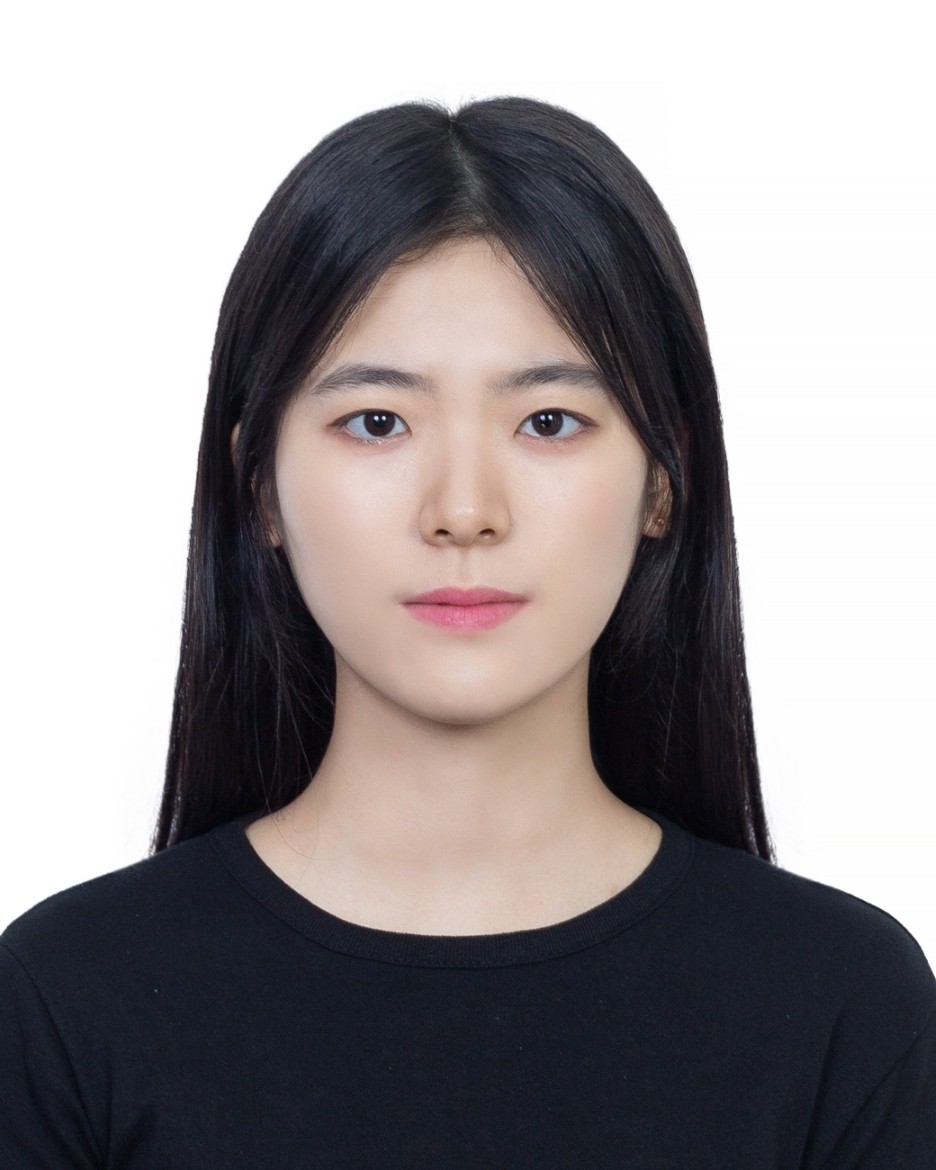}}]{EUNSEON CHOI} received the B.S. degree
in robotics from Kwangwoon University,
Seoul, South Korea, in 2025. She is currently pursuing the M.S.
degree in Convergence IT engineering from the Pohang University of Science and Technology (POSTECH). Her research interests include 
simultaneous localization and mapping, robotics,
computer vision, and robot navigation.
\end{IEEEbiography}

\begin{IEEEbiography}[{\includegraphics[width=1in,height=1.25in,clip,keepaspectratio]{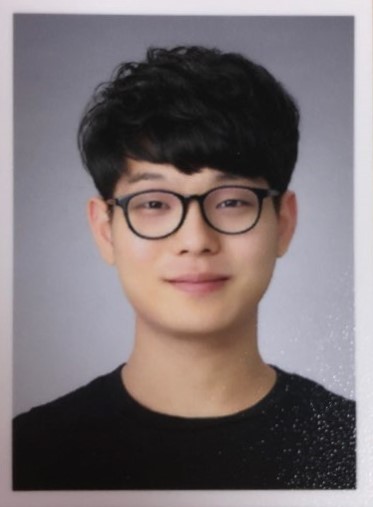}}]{JUNWOO HONG} received the B.S degree in Mechanical Engineering and M.S degree in Convergence IT Engineering from Pohang University of Science and Technology (POSTECH), in 2021 and 2023, respectively. His research interests include simultaneous localization and mapping(SLAM), robotics, robust mapping.

\end{IEEEbiography}

\begin{IEEEbiography}[{\includegraphics[width=1in,height=1.25in,clip,keepaspectratio]{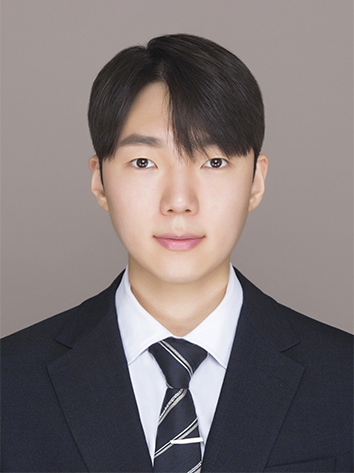}}]{DAEHAN LEE} received the B.S. degree in Automobile and IT Convergence from Kookmin University, Seoul, South Korea, in 2023. He is currently pursuing the integrated M.S. and Ph.D. degree in Convergence IT Engineering at Pohang University of Science and Technology (POSTECH), Pohang, South Korea. His research interests lie in robust state estimation and localization through multi-modal sensor fusion, supporting resilient and scalable autonomy of mobile robots in diverse and challenging environments.
\end{IEEEbiography}

\begin{IEEEbiography}[{\includegraphics[width=1in,height=1.25in,clip,keepaspectratio]{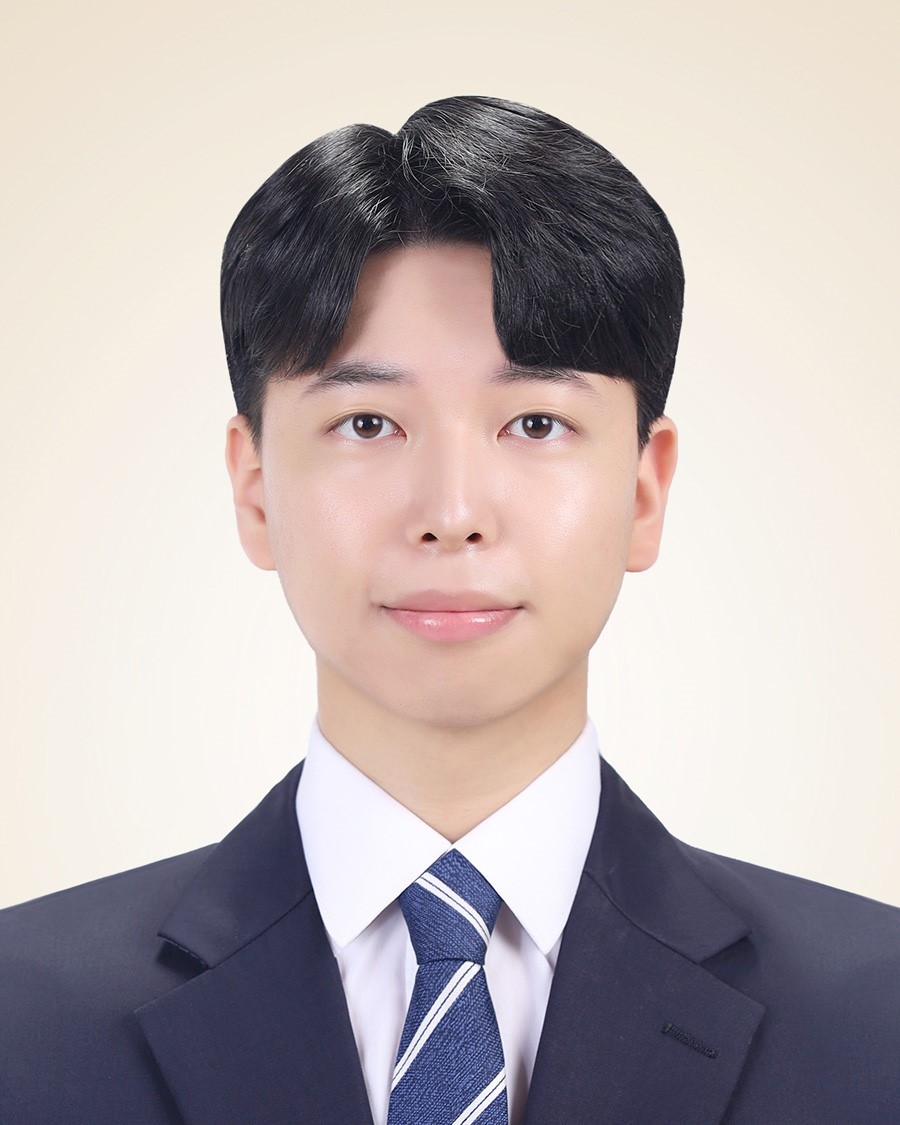}}]{sanghyun park} received the B.S. degree in School of Robotics from Kwangwoon University, Seoul, South Korea, in 2024. He is currently pursuing the Integrated-Ph.D. degree in Convergence IT Engineering. His research interests include simultaneous localization and mapping, spatial ai and robot navigation.
\end{IEEEbiography}

\begin{IEEEbiography}[{\includegraphics[width=1in,height=1.25in,clip,keepaspectratio]{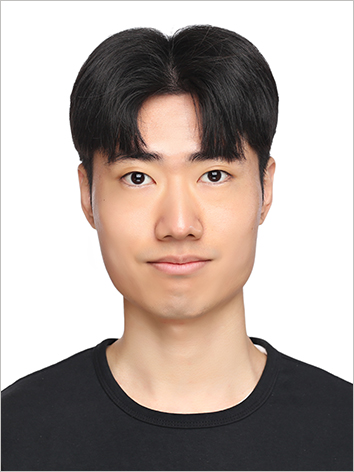}}]{HYUNYOUNG JO} received the B.S. degree in electronic engineering from the Kyungpook National University, Daegu, South Korea, in 2025. He is currently pursuing the Ph.D. degree in convergence IT engineering from the POSTECH, Pohang, South Korea. His research interests include simultaneous localization and mapping, robotics, and robot navigation.
\end{IEEEbiography}

\begin{IEEEbiography}[{\includegraphics[width=1in,height=1.25in,clip,keepaspectratio]{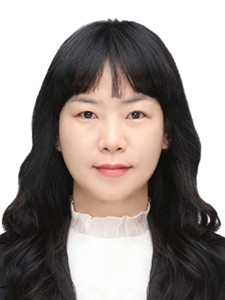}}]{SunYoung Kim} (Member, IEEE) received the B.S. degree in Electronic Engineering from Kookmin University, in 1999 and the M.S. and Ph.D. degrees in Mechanical and Aerospace Engineering from Seoul National University, in 2015 and 2019, respectively. From 2019 to 2020, she was a Postdoctoral Researcher with the School of Intelligent Mechatronics Engineering, Sejong University, Seoul, Republic of Korea. From 2020 to 2023, she was an Assistant Professor with the School of Mechanical Engineering, Kunsan National University, Jeonbuk-State, Republic of Korea. Since 2023, she has been an Associate Professor at the School of Mechanical Engineering, Kunsan National University, Jeonbuk-State, Republic of Korea. Her research interests include navigation systems, filtering, localization, GNSS interference detection and mitigation, multi-sensor fusion, multi-target tracking, target detection and classification, SLAM, autonomous vehicles, fault detection, and deep learning.
\end{IEEEbiography}

\begin{IEEEbiography}[{\includegraphics[width=1in,height=1.25in,clip,keepaspectratio]{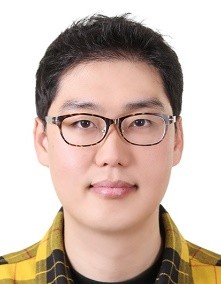}}]{ChangHo Kang} received the B.S. degree in Mechanical and Aerospace Engineering from Sejong University, in 2009, and the Ph.D. degree from the Department of Mechanical and Aerospace Engineering, Seoul National University, in 2016. From 2016 to 2018, he was a Postdoctoral Researcher with the BK21+ Transformative Training Program for Creative Mechanical and Aerospace Engineers, Seoul National University. From 2018 to 2019, he was a Research Professor with the Research Institute of Engineering and Technology, Korea University. From 2019 to 2024, he was an Assistant Professor with the Department of Mechanical System Engineering (Department of Aeronautics, Mechanical and Electronic Convergence Engineering), Kumoh National Institute of Technology. Since 2024, he has been an Assistant Professor with the Department of Artificial Intelligence and Robotics, Sejong University. His research interests include GNSS receivers, digital signal processing, nonlinear ﬁltering, and deep learning. 
\end{IEEEbiography}

\begin{IEEEbiography}[{\includegraphics[width=1in,height=1.25in,clip,keepaspectratio]{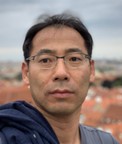}}]{SeongSam KIM} received the B.S. degree, the M.S. degree, and the Ph.D degree in Urban Engineering in 2000, 2002, and 2008, respectively from Gyeongsang National University (GNU), Jinju, Rep. of Korea. He also studied a Post-doctoral researcher at Yonsei University (Seoul, Rep. of Korea), Hong Kong Polytechnic University (Kowloon, Hong Kong), and University of Calgary (Calgary, Canada) from 2008 to 2010.
Dr. Kim is currently a senior research officer of Disaster Scientific Investigation Div. at National Disaster Management Research Institute (NDMI) of Rep. of Korea. He leads disaster and incident scene investigation research works in developing emerging and innovative technologies in combining Geomatics and state-of-art mobile observation platforms (UAVs, robot, mobile mapping system (MMS), artificial satellite include GNSS etc.) for real-time and timely response activities.
\end{IEEEbiography}

\begin{IEEEbiography}[{\includegraphics[width=1in,height=1.25in,clip,keepaspectratio]{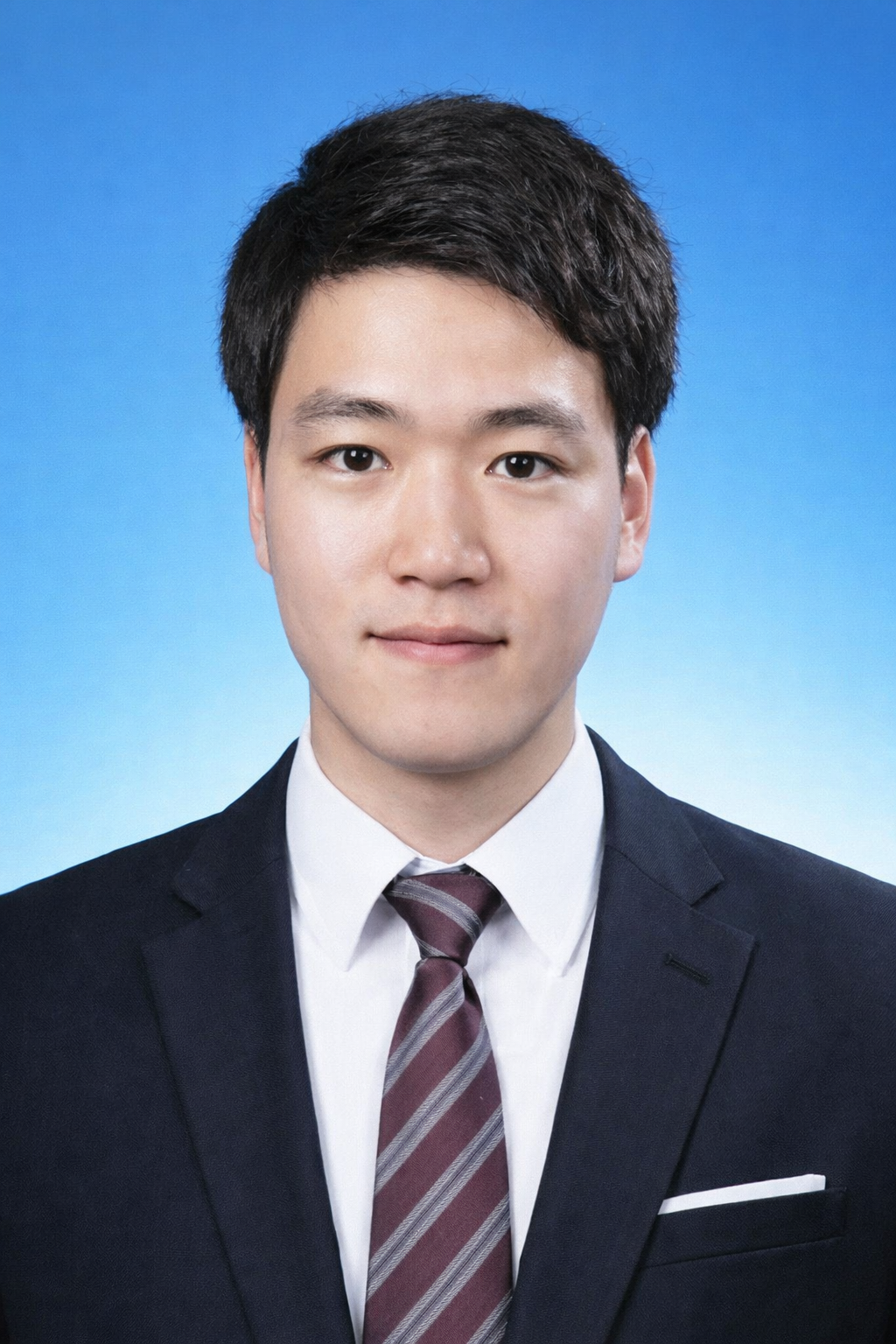}}]{YongHan JUNG} received the B.S degree, the M.S. degree in Urban Engineering in 2015, 2017, respectively from Gyeongsang National University (GNU), Jinju, Rep. of Korea. He worked at a cadastral surveying company from 2018 to 2021. He is currently a senior researcher of Disaster Scientific Investigation Div. at National Disaster Management Research Institute (NDMI) of Rep. of Korea. His main research interests include disaster investigation and analysis using UAVs and LiDAR.
\end{IEEEbiography}

\begin{IEEEbiography}[{\includegraphics[width=1in,height=1.25in,clip,keepaspectratio]{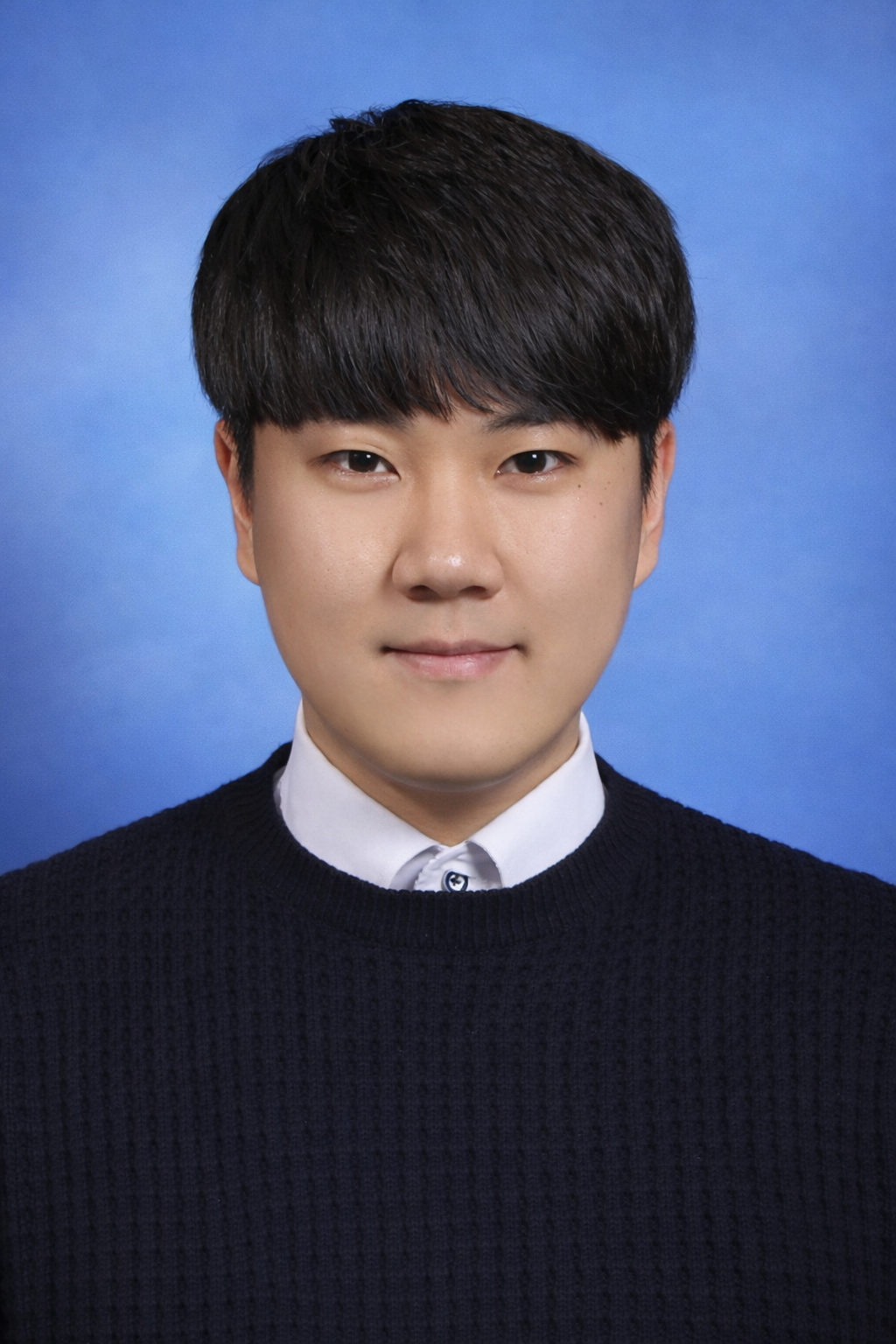}}]{JungWook PARK} received the B.S. and M.S. degrees in Architecture and Fire Safety from Dongyang University, Yeongju, Rep. of Korea, in 2019 and 2022. He is currently a researcher of Disaster Scientific Investigation Div. at National Disaster Management Research Institute (NDMI) of Rep. of Korea. His research interests include UAV-based mapping, LiDAR data processing, and three-dimensional modelling of disaster sites with applications to disaster investigation and risk assessment.
\end{IEEEbiography}

\begin{IEEEbiography}[{\includegraphics[width=1in,height=1.25in,clip,keepaspectratio]{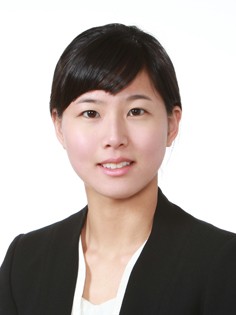}}]{Seul KOO} received the B.S degree, the M.S. degree in Urban Engineering in 2011, 2013, respectively from Gyeongsang National University (GNU), Jinju, Rep. of Korea. She is currently a researcher of Disaster Scientific Investigation Div. at National Disaster Management Research Institute (NDMI) of Rep. of Korea. Her research interests include GIS, LiDAR data processing with applications to disaster investigation, and risk assessment.
\end{IEEEbiography}

\begin{IEEEbiography}[{\includegraphics[width=1in,height=1.25in,clip,keepaspectratio]{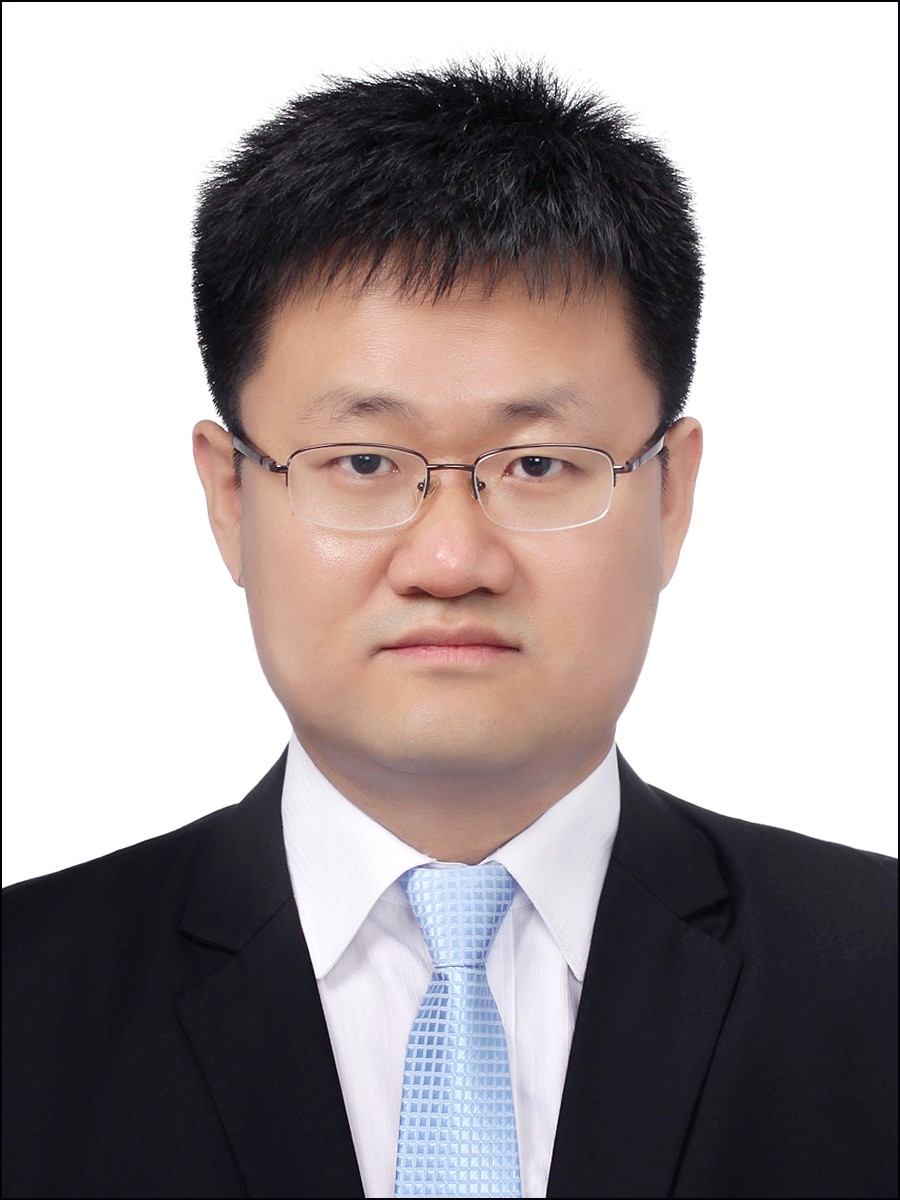}}]{SOOHEE HAN} (Senior Member, IEEE) received the B.S. degree in electrical engineering in
1998, and the M.S. and Ph.D. degrees in electrical engineering and computer science in 2000
and 2003, respectively, all from Seoul National University (SNU), Seoul, South Korea.
From 2003 to 2007, he was a Researcher with the Engineering Research Center for Advanced
Control and Instrumentation, SNU. In 2005, he was a Visiting Scholar with Stanford University.
From 2009 to 2014, he was with the Department of Electrical Engineering, Konkuk University, Seoul,
South Korea. Since 2014, he has been with the Department of Electrical
Engineering and Convergence IT Engineering, Pohang University of Science and Technology, Pohang, South Korea. In 2023, he was a Visiting Professor with the University of Waterloo, Waterloo, Ontario, Canada. His main research interests include real-time reinforcement learning, mathematical instrumentation, and model-based battery informatics.

Dr. Han is currently a Technical Editor for the {\it{IEEE Transactions on Mechatronics}}, 
and the {\it{Journal of Electrical Engineering and Technology}}. In 2019, he served as an Editor for {\it{IFAC Workshop on Control of Smart Grid and Renewable Energy Systems.}}
\end{IEEEbiography}

\EOD

\end{document}